\documentclass[twoside]{article}
\usepackage{cite}
\usepackage{amsfonts,amssymb,amsbsy,textcomp,marvosym,picins,amsmath,caption,threeparttable,amsthm}
\usepackage{eurosym,mathrsfs,fancyhdr,graphics,indentfirst,color,bm,upgreek,booktabs,graphicx}
\usepackage{multicol}
\usepackage{CJK}
\usepackage{framed,multirow}

\usepackage{amssymb}
\usepackage{latexsym}
\usepackage{url}
\usepackage{xcolor}
\definecolor{newcolor}{rgb}{.8,.349,.1}

\usepackage{epsfig}
\usepackage{amssymb}
\usepackage{amsmath}
\usepackage{amsthm}
\usepackage{cases}
\usepackage{esint}
\usepackage{graphicx}
\usepackage{subfig}
\usepackage{algorithm} 
\usepackage{algorithmic} 
\usepackage{multirow} 
\usepackage{amsmath}
\usepackage{xcolor}
\usepackage{gensymb}

\usepackage{makecell}
\usepackage{overpic}
\usepackage{multirow}
\usepackage{multicol}

\usepackage{overpic}
\newcommand{\sign}[1]{(-1)^{\mathbb{I}_{#1<0}}}
\newcommand{\signtxt}[0]{\text{sign}}
\newcommand{\abs}[0]{\text{abs}}
\newcommand{\edge}[1]{\langle #1 \rangle}
\newcommand{\floor}[1]{\lfloor #1 \rfloor}
\newcommand{\round}[1]{\lfloor #1+\frac{1}{2} \rfloor}
\newcommand{\clip}[3]{\lfloor #1 \rceil_{#2}^{#3}}
\newcommand{\gcc}[1]{#1}
\newcommand{\gccpf}[1]{#1}
\newcommand{\cumparagraph}[1]{\noindent\gccpf{\textbf{#1}}}
\newcommand{\textnbf}[1]{#1}

\newcommand{\secref}[1]{\gccpf{Section~\ref{#1}}}
\newcommand{\subsecref}[1]{\gccpf{Subsection~\ref{#1}}}
\newcommand{\figref}[1]{Fig.~\ref{#1}}

\newcommand{\tabref}[1]{Table~\ref{#1}}
\newcommand{\algref}[1]{\gccpf{Algorithm~\ref{#1}}}
\newcommand{\sArt}[0]{state-of-the-art}

\usepackage{ragged2e}
\def\footnoterule{\kern 1mm \hrule width 10cm \kern 2mm}

\allowdisplaybreaks
\catcode`@=11
\def\title#1{\vspace{3mm}\begin{flushleft}\vglue-.1cm\Large\bf\boldmath\protect\baselineskip=18pt plus.2pt minus.1pt #1
\end{flushleft}\vspace{1mm} }
\def\author#1{\begin{flushleft}\normalsize #1\end{flushleft}\vspace*{-4pt} \vspace{3mm}}
\def\address#1#2{\begin{flushleft}\vglue-.35cm${}^{#1}$\small\it #2\vglue-.35cm\end{flushleft}\vspace{-2mm}\par}

\catcode`@=11
\def\section{\@startsection{section}{1}{\z@}%
 {-3ex \@plus -.3ex \@minus -.2ex}%
 {2.2ex \@plus.2ex}%
{\normalfont\normalsize\protect\baselineskip=14.5pt plus.2pt minus.2pt\bfseries}}
\def\subsection{\@startsection{subsection}{2}{\z@}%
 {-3ex\@plus -.2ex \@minus -.2ex}%
 {2ex \@plus.2ex}%
{\normalfont\normalsize\protect\baselineskip=12.5pt plus.2pt minus.2pt\bfseries}}
\def\subsubsection{\@startsection{subsubsection}{3}{\z@}%
 {-2.2ex\@plus -.21ex \@minus -.2ex}%
 {1.4ex \@plus.2ex}
{\normalfont\normalsize\protect\baselineskip=12pt plus.2pt minus.2pt\sl}}

\begin{document}
\begin{CJK*}{UTF8}{gbsn}
\thispagestyle{empty}
\vspace*{-13mm}
\noindent {\small Cheng Gong, Ye Lu, Surong Dai {\it et al.} AutoQNN: An End-to-End Framework for Automatically Quantizing Neural Networks.
JOURNAL OF COMPUTER SCIENCE AND TECHNOLOGY \ 33(1): \thepage--\pageref{last-page}
\ January 2018. DOI 10.1007/s11390-015-0000-0}
\vspace*{2mm}

\title{AutoQNN: An End-to-End Framework for Automatically Quantizing Neural Networks}

\author{Cheng Gong$^{1}$ (龚~~成), \emph{(Student) Member}, \emph{CCF}, Ye Lu$^{2,3}$ (卢~~冶), \emph{(Senior) Member}, \emph{CCF}, Su-Rong Dai$^{2}$ (代素蓉), \emph{(Student) Member}, \emph{CCF}, Qian Deng$^{2}$ (邓~~倩), Cheng-Kun Du$^{2}$ (杜承昆), \emph{(Student) Member}, \emph{CCF}, and Tao Li$^{2,3,*}$ (李~~涛), \emph{(Distinguished) Member}, \emph{CCF}, \emph{Member}, \emph{ACM}}

\address{1}{College of Software, Nankai University, Tianjin 300350, China}

\address{2}{College of Computer Science, Nankai University, Tianjin 300350, China}

\address{3}{State Key Laboratory of Computer Architecture, Institute of Computing Technology, Chinese Academy of Sciences, Beijing 100190, China}

\vspace{2mm}

\noindent E-mail: cheng-gong@nankai.edu.cn; luye@nankai.edu.cn; daisurong@mail.nankai.edu.cn; dengqian@mail.nankai.edu.cn; dck@mail.nankai.edu.cn; litao@nankai.edu.cn
\\[-1mm]

\noindent Received May 30, 2021.\\[1mm]

\let\thefootnote\relax\footnotetext{{}\\[-4mm]\indent\ Regular Paper\\[.5mm]
\indent\ This work is partially supported by
the National Key Research and Development Program of China under Grant No.~2018YFB2100300,
the National Natural Science Foundation under Grant No.~62002175 and No.~62272248,
the State Key Laboratory of Computer Architecture at the Institute of Computing Technology, Chinese Academy of Sciences, under Grant No.~CARCHB202016 and No.~CARCHA202108,
the Special Funding for Excellent Enterprise Technology Correspondent of Tianjin under Grant No.~21YDTPJC00380,
the Open Project Foundation of Information Security Evaluation Center of Civil Aviation, Civil Aviation University of China under Grant No.~ISECCA-202102, 
and the CCAI-Huawei MindSpore Open Fund under Grant No. CAAIXSJLJJ-2021-025A. \\[.5mm]
\indent\ $^*$Corresponding Author
\\[.5mm]\indent\ \copyright Institute of Computing Technology, Chinese Academy of Sciences 2021}

\noindent{\small\bf Abstract} \quad  {\small 
\justifying
Exploring the expected quantizing scheme with suitable mixed-precision policy is the key point to compress deep neural networks (DNNs) in high efficiency and accuracy. This exploration implies heavy workloads for domain experts, and an automatic compression method is needed.
However, the huge search space of the automatic method introduces plenty of computing budgets that make the automatic process challenging to be applied in real scenarios.
In this paper, we propose an end-to-end framework named AutoQNN, for automatically quantizing different layers utilizing different schemes and bitwidths without any human labor.
AutoQNN can seek desirable quantizing schemes and mixed-precision policies for mainstream DNN models efficiently by involving three techniques: quantizing scheme search (QSS), quantizing precision learning (QPL), and quantized architecture generation (QAG).
QSS introduces five quantizing schemes and defines three new schemes as a candidate set for scheme search, and then uses the differentiable neural architecture search (DNAS) algorithm to seek the layer- or model-desired scheme from the set.
QPL is the first method to learn mixed-precision policies by reparameterizing the bitwidths of quantizing schemes, to the best of our knowledge. QPL optimizes both classification loss and precision loss of DNNs efficiently and obtains the relatively optimal mixed-precision model within limited model size and memory footprint.
QAG is designed to convert arbitrary architectures into corresponding quantized ones without manual intervention, to facilitate end-to-end neural network quantization.
We have implemented AutoQNN and integrated it into Keras.
Extensive experiments demonstrate that AutoQNN can consistently outperform state-of-the-art quantization.  
For \text{2-bit} weight and activation of AlexNet and ResNet18, AutoQNN can achieve the accuracy results of \text{59.75\%} and \text{68.86\%}, respectively, and obtain accuracy improvements by up to \text{1.65\%} and \text{1.74\%}, respectively, compared with state-of-the-art methods. 
Especially, compared with the full-precision AlexNet and ResNet18, the \text{2-bit} models only slightly incur accuracy degradation by \text{0.26\%} and \text{0.76\%}, respectively, which can fulfill practical application demands.

}

\vspace*{3mm}

\vspace*{4mm}

\end{CJK*}
\baselineskip=18pt plus.2pt minus.2pt
\parskip=0pt plus.2pt minus0.2pt
\begin{multicols}{2}
\section{Introduction}
The heavy computational burden immensely hinders the deployment of deep neural networks (DNNs) on resource-limited devices in real application scenarios.
Quantization is a technique which compresses DNN weight and activation values from high-precision to low-precision.
\gcc{The low-precision weights and activation occupy smaller memory bandwidth, registers, and computing units, thus significantly improving the computing performance. For example, AlexNet with 32-bit floating-point (FP32) can only achieve the performance of 1.93TFLOPS on RTX2080Ti since the bandwidth constraints.
However, the low-precision AlexNet with 16-bit floating-point (FP16) can achieve the performance of 7.74TFLOPS on the same device because the required bandwidth is halved while the available computing units are doubled\footnote{The results are computed with the Roofline algorithm~\cite{williams2009roofline} using the memory access quantity and operations of an official AlexNet model from the PyTorch community (https://github.com/pytorch/vision) and the peak performance and memory bandwidth of RTX2080Ti GPU.}. The FP16 AlexNet is four times faster than the FP32 one on RTX2080Ti GPU.}
\gcc{Therefore, quantization} can reduce the computing budgets in DNN inference phases and enable DNN model deployment on resource-limited devices.

However, unreasonable quantizing strategies, such as binary~\cite{hubara2016binarized} and ternary~\cite{TWNs}, tend to seriously affect DNN model accuracy~\cite{zhou2016dorefa,Ternaryconnect} and lead to customer frustration.
\gcc{Lower bitwidth usually leads to higher computing performance but larger accuracy degradation.
The quantization strategy selection dominates the computing performance and inference accuracy of models.
In order to balance the computing performance and inference accuracy, many previous \gccpf{investigations} have tried to select a unified bitwidth for all layers in DNNs \cite{VecQ,cheng2019uL2Q} cautiously. 
However, many \gccpf{studies} show that different layers of DNNs have different sensitivities~\cite{HAWQ,HAWQV2} and computing budgets~\cite{wang2019haq}.
Using the same bitwidth for different layers is hard to obtain superior speed-up and accuracy.}
To strike a fine-grained balance between efficiency and accuracy, it is strongly demanded to explore desirable quantizing schemes~\cite{lin2016fixed} and reasonable mixed-precision policies~\cite{wang2019haq,wu2018mixed} for various neural architectures.
The brute force approaches are not feasible for this exploration since the search space grows exponentially with the number of layers~\cite{HAWQ}. 
Heuristic explorations work in some scenarios but greatly rely on the heavy workloads of domain experts~\cite{lin2016fixed}. 
Besides, it is impractical to find desirable strategies for various neural architectures through manual-participated heuristic explorations because there are a large number of different neural architectures, and the number of neural architectures is still explosively increasing.

Therefore, it is expected to propose an automatic DNN quantization without manual intervention.
The challenge of automatic quantization lies in efficiently exploring \gccpf{the large search space that exponentially increases with the number of layers in DNNs}.
Many \gccpf{studies} have made substantial efforts and gained tremendous advances in this area, \gccpf{such as mixed-precision (MixedP)~\cite{wu2018mixed}, HAQ~\cite{wang2019haq}, AutoQB~\cite{AutoQB}, and HAWQ~\cite{HAWQ}}.
Nevertheless, there are still some issues that have not been resolved well yet, as follows.
Firstly, it is widely acknowledged that different quantizing schemes can impose an impact on the accuracy of quantized DNNs even with the same quantizing bitwidth~\cite{zhu2016TTQ,zhou2016dorefa,ENN2017}.
\gccpf{However, few studies investigate seeking quantizing schemes for specific architectures, to our best knowledge.}
Secondly, mixed-precision quantization is an efficient way to improve the accuracy of quantized DNNs without increasing the average bitwidth~\cite{wang2019haq,wu2018mixed,AutoQB}.
\gcc{The presented algorithms include reinforcement learning~\cite{wang2019haq,AutoQB}, evolution-based search~\cite{BMobileES}, and hessian-aware methods~\cite{HAWQ,HAWQV2}. 
However, these algorithms for mixed-precision search are highly complex and inefficient in exploring the exponential search space.}
The algorithms usually require lots of computing resources, making it challenging to deploy them in online learning scenarios.
Besides, these algorithms can easily fall into sub-optimal solutions because they usually skip the search steps of unusual bitwidths, such as a bitwidth of 5~\cite{wu2018mixed}, to reduce search time.

To address the above issues and challenges, we propose AutoQNN, an end-to-end framework for automatically quantizing neural networks without manual intervention. 
AutoQNN seeks desirable quantizing strategies for arbitrary architectures by involving three techniques: quantizing scheme search (QSS), quantizing precision learning (QPL), and quantized architecture generation (QAG).
The extensive experiments demonstrate that AutoQNN has the ability to find expected quantizing strategies within finite time and consistently outperforms the~\sArt~quantization methods for various models on ImageNet.
Our contributions are summarised as follows.
\begin{itemize}
\item 
\gccpf{We propose QSS to automatically find desirable quantizing schemes for weights and activation with various distributions.} 

\item 
\gcc{\gccpf{We present QPL to efficiently learn the relatively optimal mixed-precision policies by minimizing the classification and precision losses.}
}

\item 
\gccpf{We design QAG to automatically convert the computing graphs of arbitrary DNNs into quantized architectures without manual intervention.}
\item \gccpf{Extensive experiments highlight that AutoQNN can find the expected quantizing strategies to reduce accuracy degradation with low bitwidth.} 

\end{itemize}

The rest of this paper is organized as follows. 
We first discuss the related work in Section 2 and then elaborate QSS and QPL in Section 3 and Section 4, respectively. 
Next, we introduce QAG and end-to-end framework in Section 5, and evaluate AutoQNN in Section 6. 
Finally, we conclude this paper in Section 7.

\section{Related \gccpf{work}}\label{sec:relatedworks}
Quantization has been deeply investigated as an efficient approach to \gccpf{boosting} DNN computing efficiency. In this section, we first introduce the related quantizing schemes, and then describe the recent mixed-precision quantizing strategies.

\subsection{Quantizing Schemes}
Binary \gccpf{methods} with only one bit, such as BC~\cite{Binaryconnect}, BNN~\cite{hubara2016binarized}, and Xnor-Net~\cite{rastegari2016xnor}, prefer noticeable efficiency improvement.
They can compress DNN memory footprint by up to \gccpf{32x} and replace expensive multiplication with cheap bit-operations.
However, the binary \gccpf{methods} significantly degrade accuracy, since the low bitwidth loses much information. 
Ternary \gccpf{methods}~\cite{TWNs,alemdar2017ternary,Ternaryconnect,jin2018sparse} quantize the weights or activation of DNNs into ternaries of \gccpf{\{-1, 0, 1\}}, aiming at remedying the accuracy degradation of the binary \gccpf{methods} and not introducing additional overheads. 
\gccpf{Quaternary methods~\cite{cheng2019uL2Q,VecQ} quantize model weights into four values of \{-2, -1, 0, 1\}. They reduce the model accuracy degradation using the same two bits as the ternary methods.}

Despite the advanced computing efficiency of the 1-bit or 2-bit methods above, the low bitwidth can significantly affect the model accuracy. 
This motivates researchers to investigate high bitwidth fixed-point quantization.
The proposed method in~\cite{gysel2016hardware} abandons the last bits of the binary strings of values and keeps the remaining bits as quantized fixed-point values. 
T-DLA \cite{chen2019tdla} quantizes values into low-precision fixed-point values by reserving the first several significant bits of the binary string of values and dropping others. 
However, it is challenging for fixed-point quantization to handle the weights with a high dynamic range.

Zoom quantization methods handle the weights with a high dynamic range by multiplying a full-precision scaling factor.
For example,
Dorefa-Net \cite{zhou2016dorefa} and STC~\cite{jin2018sparse} first zoom the weights into the range of [-1,1] by dividing the weights according to their maximum value, and then uniformly map them into continuous integers. 
\gccpf{Zoom} quantization does not deal with the outliers in weights, which increases the quantization loss.

\gccpf{Clip quantization eliminates the impact of outliers on zoom methods by estimating a reasonable range. It truncates weights into the estimated range.} 
The key point of clip quantization is how to balance clip-errors and project-errors~\gccpf{\cite{APoT,TSQ2018,QIL}}.
PACT~\cite{PACT} reparameterizes the clip threshold to learn a reasonable quantizing range.
HAQ~\cite{wang2019haq} minimizes the KL-divergence between the original weight distribution and the quantized weight distribution to seek the optimal clip threshold.
$\mu$L2Q~\cite{cheng2019uL2Q} seeks the optimal quantizing \gccpf{step-size} by minimizing the L2 distance between the original values and the quantized ones.

Besides, there are many studies concerning non-uniform quantization, \gccpf{such as power-of-two (PoT) quantization and residual quantization}.
\gccpf{PoT methods~\cite{LogQ,INQ2017,ENN2017} quantize values into the form of PoT thus converting the multiplication into addition~\cite{LogQ}.}
Residual \gccpf{methods}~\cite{ghasemzadeh2018rebnet,Rebinary,he2016resnet,residualquantization2021tom} quantize the residual errors, which are produced by the last quantizing process, into binaries iteratively. 
Similar \gccpf{methods}, such as LQ-Nets \cite{LQ-Nets}, ABC-Net \cite{ABC-Net}, and AutoQ~\cite{AutoQ}, quantize the weights/activation into the sum of several binary results.

\subsection{Mixed-Precision Strategies}       
Mixed precision is well known as an efficient quantizing strategy, which quantizes different layers with different bitwidths, thus achieving high accuracy with low average quantizing bitwidth. 
HAQ~\cite{wang2019haq} leverages reinforcement learning (RL) to automatically determine the bitwidth of layers of DNNs by receiving the hardware accelerator's feedback in the design loop.
Mixed-precision method (MixedP)~\cite{wu2018mixed} formulates mixed-precision quantization as a neural architecture search problem. 
AutoQB~\cite{AutoQB} introduces deep reinforcement learning (DRL) to automatically explore the space of fine-grained channel-level network quantization. 
The proposed method in~\cite{BMobileES} converts each depth-wise convolution layer in MobileNet to several group convolution layers with binarized weights, \gccpf{and employs the evolution-based search to explore the number of group convolution layers.} 
\gccpf{HAWQ~\cite{HAWQ} and HAWQ-V2~~\cite{HAWQV2}} employ the second-order information, i.e., top Hessian eigenvalue and Hessian trace of weights/activation, to compute the sensitivities of layers and then design a mixed-precision policy based on these sensitivities.
\gccpf{BSQ~\cite{yang2021bsq} considers each bit of the quantized weights as an independent trainable variable and introduces a differentiable bit-sparsity regularizer for reducing precision.
}
\section{Quantizing Scheme Search}\label{sec:qss}
In this section, we elaborate the automatic quantizing scheme search. 
We first summarize \gcc{five} classical quantizing schemes, then propose \gcc{three} new quantizing schemes, and finally take the eight schemes as a candidate set. 
We will seek desirable schemes from the candidate set for arbitrary architectures. 
For ease of notation, we define $\gccpf{\boldsymbol{d_\text{f}}}\in \mathbb{R}^d$ and $\gccpf{\boldsymbol{d_\text{q}}}\in\mathbb{Q}^d$ as the original and quantized vectors, respectively. $\mathbb{R}$ is real space and $\mathbb{Q}$ is the set of quantized values.

\begin{figure*}[!ht]
    \centering
    \subfloat[]{
    \label{fig:zoomq-figures}
    \includegraphics[width=0.33\textwidth]{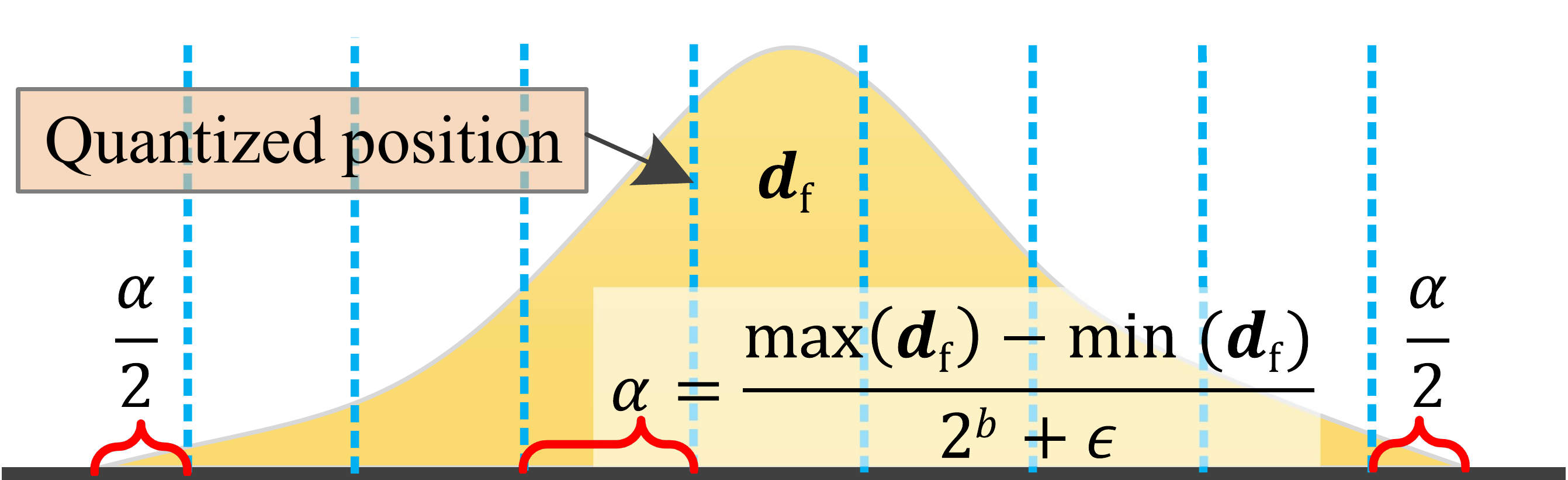}}
    \subfloat[]{
    \label{fig:clipq-figures}
    \includegraphics[width=0.33\textwidth]{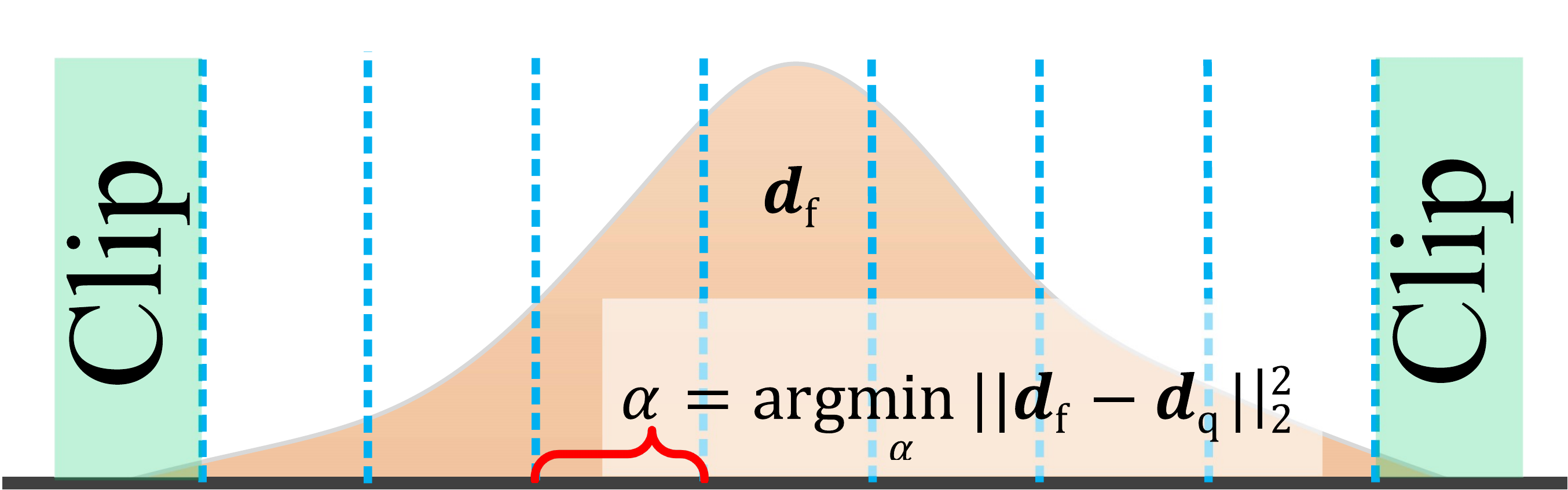}}
    \subfloat[]{
    \label{fig:potq-figures}
    \includegraphics[width=0.33\textwidth]{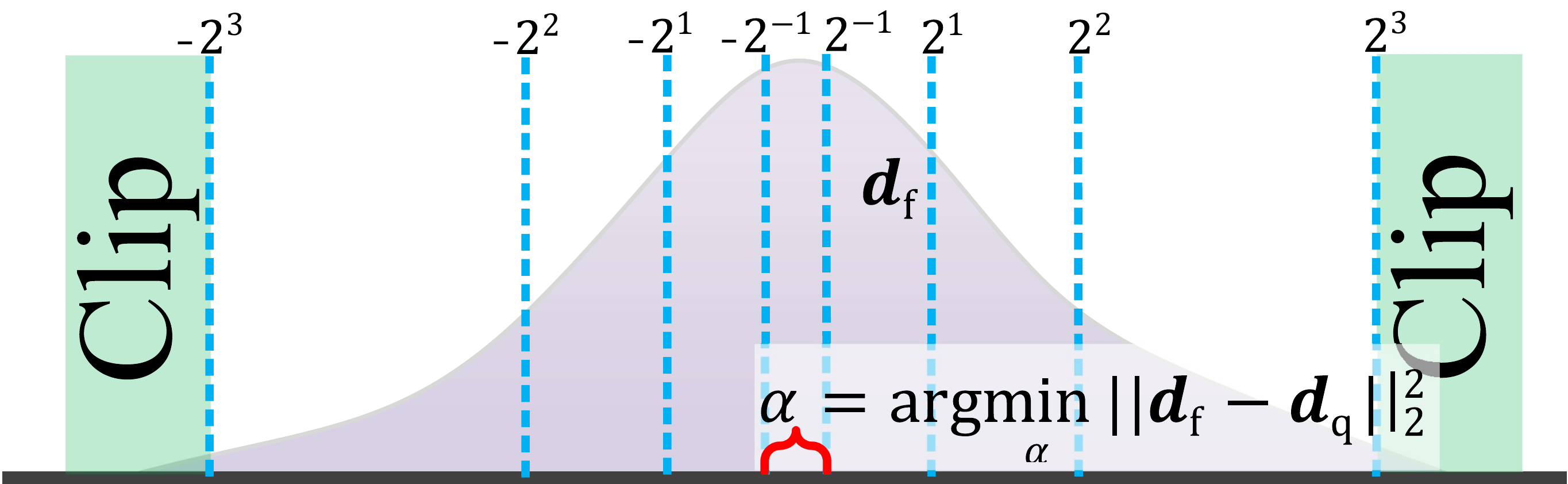}}
    \caption{The quantizing process of our proposed new schemes, including \gccpf{(a) zoom quantization, (b) clip quantization, and (c) PoT quantization.}}
    \label{fig:QPL_on_cifar10}
\end{figure*}
\subsection{Candidate Schemes}\label{sec:candidate_schemes}
We divide related quantization methods into eight categories based on their quantizing processes and the format of quantized values: \textnbf{Binary}, \textnbf{Ternary}, \textnbf{Quaternary}, \gccpf{fixed} quantization (\textnbf{FixedQ}), \gccpf{residual} quantization (\textnbf{ResQ}), \gccpf{zoom} quantization (\textnbf{ZoomQ}), \gccpf{clip} quantization (\textnbf{ClipQ}), and PoT quantization (\textnbf{PotQ}). 

The first five schemes can be realized easily, so we just refer to the existing methods.

\noindent\textnbf{Binary.} The binary in~\cite{rastegari2016xnor} is summarized as follows:
\begin{equation}
    \begin{split}
        \gccpf{\boldsymbol{d_\text{q}}}=\alpha\cdot \signtxt(\gccpf{\boldsymbol{d_\text{f}}}) 
        ~~\text{s.t.}~\alpha=E(\abs(\gccpf{\boldsymbol{d_\text{f}}})).
    \end{split}\nonumber
\end{equation}
Here $\signtxt(\gccpf{\boldsymbol{d_\text{f}}})=\sign{\gccpf{\boldsymbol{d_\text{f}}}}$ maps the positive elements of $\gccpf{\boldsymbol{d_\text{f}}}$ to $1$ and non-positive ones to \gccpf{$-1$}. 
$\abs(\gccpf{\boldsymbol{d_\text{f}}})=\signtxt(\gccpf{\boldsymbol{d_\text{f}}})\cdot \gccpf{\boldsymbol{d_\text{f}}}$ converts the elements of $\gccpf{\boldsymbol{d_\text{f}}}$ into their absolute values. $E(\cdot)$ computes the expectation of input. 

\noindent\textnbf{Ternary.} The ternary proposed in TWN~\cite{TWNs} is reorganized as follows:
\begin{equation}
\begin{split}
    \gccpf{\boldsymbol{d_\text{q}}}&=\alpha\cdot \clip{\round{\frac{\gccpf{\boldsymbol{d_\text{f}}}}{2\beta}}}{-1}{1}\\
    &\text{s.t.}~\alpha=E(\abs(\{x|x\in \gccpf{\boldsymbol{d_\text{f}}},x>\beta\})),\\
    &~~~~~\beta=0.7E(\abs(\gccpf{\boldsymbol{d_\text{f}}})).
\end{split}\nonumber
\end{equation}
Here $\lfloor\cdot\rfloor$ is the floor function, and $\clip{\cdot}{-1}{1}$ truncates the elements of a vector to the range of $[-1,$ $1]$. 

\noindent\textnbf{Quaternary.} Quaternary quantizes weights into four values of $\{-2,$ $-1,$ $0,$ $1\}$. The quaternary definition in \cite{VecQ}~is summarized as follows:
\begin{equation}
    \begin{split}
        \gccpf{\boldsymbol{d_\text{q}}}&=\alpha\cdot(\clip{\floor{\frac{\gccpf{\boldsymbol{d_\text{f}}}}{\alpha}}}{-2}{1}+\frac{1}{2})\\
        &\text{s.t.}~\alpha=\sqrt{D(\gccpf{\boldsymbol{d_\text{f}}})}.
    \end{split}\nonumber
\end{equation}
Here $D(\cdot)$ computes the variance of input.

\noindent\textnbf{FixedQ.} FixedQ quantizes values into low-precision fixed-point formats by dropping several bits of the binary strings of values. 
The FixedQ in~\cite{chen2019tdla} is summarized as follows:
\begin{equation}
    \begin{split}
        \gccpf{\boldsymbol{d_\text{q}}}&=\alpha\cdot\round{\frac{\gccpf{\boldsymbol{d_\text{f}}}}{\alpha}}\\
        &\text{s.t.}~\alpha=2^p,p=\floor{\log_2(\max(\abs(\gccpf{\boldsymbol{d_\text{f}}})))}-(b-2).
    \end{split}\nonumber
\end{equation}
Here $b$ is bitwidth. 
$\max(\cdot)$ finds the maximum element of an input vector. $\log_2(\cdot)$ calculates the logarithm of a scalar with the base of two.

\noindent\textnbf{ResQ.}
ResQ quantizes the residual errors, which are the quantization errors produced by the last quantizing process, into binaries iteratively. 
ResQ can be defined as follows:
\begin{equation}
\small
\begin{split}
    \gccpf{\boldsymbol{d_\text{q}}}&=\sum_{i=1}^{b} B(\boldsymbol{v}_i)\\
    &\text{s.t.}~B(\boldsymbol{v})=E(\abs(\boldsymbol{v}))\cdot \signtxt(\boldsymbol{v}),\\
    &~~~~~\boldsymbol{v}_i=
    \begin{cases}
    \gccpf{\boldsymbol{d_\text{f}}}, ~\text{if}~i=1\\
    \boldsymbol{v}_{i-1}-B(\boldsymbol{v}_{i-1}),~\text{if}~i=2,3,\cdots,b.
    \end{cases}
\end{split}\nonumber
\end{equation}

The remaining schemes, including \textnbf{ZoomQ}, \textnbf{ClipQ}, and \textnbf{PotQ}, are widely applied and have been realized in a variety of ways. Here we propose \textnbf{three} new definitions of them below.

\begin{figure*}
    \centering
    \includegraphics[width=1\textwidth]{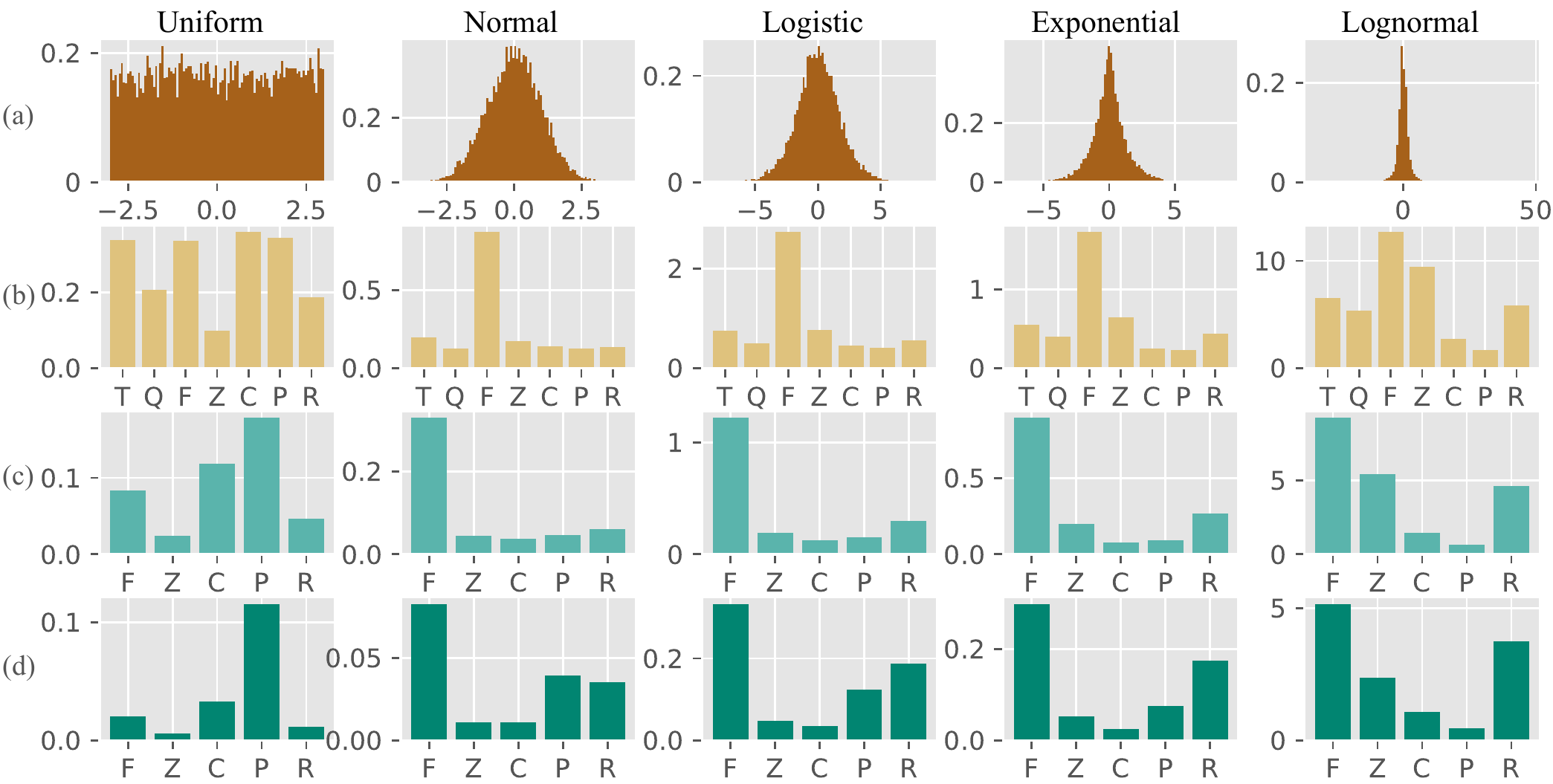}
    \vspace{-20pt}
    \caption{Quantization loss comparisons of candidate schemes across various data distribution and quantizing bits. The candidate schemes include Ternary (T), Quaternary (Q), \gcc{FixedQ} (F), \gcc{ZoomQ} (Z), \gcc{ClipQ} (C), \gcc{PotQ} (P) and \gcc{ResQ} (R). Quantization loss is the L2 distance between values before and after quantization.
    \gccpf{(a) Different data distributions. (b) 2-bit quantization losses. (c) 3-bit quantization losses. (d) 4-bit quantization losses.}}
    \label{fig:Qloss_vs_schemes_vs_bits}
    \vspace{-5pt}
\end{figure*}
\noindent\textnbf{ZoomQ}. ZoomQ indicates a group of schemes that uniformly map full-precision values into integers. 
ZoomQ is usually realized by zooming and rounding operations. The new definition of ZoomQ is given below.
\begin{equation}
\small
    \begin{split}
        \gccpf{\boldsymbol{d_\text{q}}}&=\alpha\cdot\clip{\floor{\frac{\gccpf{\boldsymbol{d_\text{f}}}-\beta}{\alpha}}}{0}{2^{b}-1}+\beta+\frac{\alpha}{2}\\
        &\text{s.t.}~\alpha=\frac{\max(\gccpf{\boldsymbol{d_\text{f}}})-\min(\gccpf{\boldsymbol{d_\text{f}}})}{2^{b}},~\beta=\min(\gccpf{\boldsymbol{d_\text{f}}}).
    \end{split}\nonumber
\end{equation}
Here $\min(\cdot)$ returns the minimum element of an input vector. The quantizing process of ZoomQ is shown in \gccpf{\figref{fig:QPL_on_cifar10}(a)}. We first compute a quantizing interval width $\alpha$, and then map the full precision $\gccpf{\boldsymbol{d_\text{f}}}$ into the integer vector based on the obtained $\alpha$.

\noindent\textnbf{ClipQ}. ClipQ first truncates values into a target range, and then uniformly maps the values to low-precision representations. 
We define ClipQ as follows:
\begin{equation}
\small
    \begin{split}
        \gccpf{\boldsymbol{d_\text{q}}}&=\alpha\cdot(\clip{\floor{\frac{\gccpf{\boldsymbol{d_\text{f}}}}{\alpha}}}{-2^{b-1}}{2^{b-1}-1}+\frac{1}{2})\\
        &\text{s.t.}~\alpha=\underset{\alpha}{\arg\min}~||\gccpf{\boldsymbol{d_\text{f}}}-\gccpf{\boldsymbol{d_\text{q}}}||_2^2.
    \end{split}\nonumber
\end{equation}
The quantizing process of ClipQ is drawn in \gccpf{\figref{fig:QPL_on_cifar10}(b)}.
For simplicity, we compute optimal $\alpha$ for each quantizing bitwidth offline by assuming that $\boldsymbol{d_\text{f}}$ satisfies normal distribution~\cite{cheng2019uL2Q,VecQ}. 
The optimal solutions under different bitwidths are $\alpha\in\{1.2832,$ $0.6694,$ $0.3570,$ $0.1939,$ $0.1056,$ $0.0573,$ $0.0308\}$ for $b\in \{2,$ $3,$ $4,$ $5,$ $6,$ $7,$ $8\}$.

\renewcommand{\arraystretch}{1.0}
\begin{table*}[!t]
\centering
\caption{\gccpf{Detailed Information of the Eight Candidate Schemes}\label{tab:candidate_scheme_table}}
\vspace{-5pt}
\setlength{\tabcolsep}{11pt}
\begin{tabular}{lccccccc}\Xhline{1.5pt}
Schemes &
  \gccpf{Number of Bits} &
  TC&
  SC&
  Distributions&
  Formats &
  \gccpf{Reference}  \\\Xhline{.5pt}
Binary     & 1   & $O(n)$      & $O(n)$      & - & Binary & \cite{Binaryconnect,hubara2016binarized,rastegari2016xnor} \\
Ternary    & 2   & $O(n)$      & $O(n)$      & Normal                      & Ternary& \cite{TWNs,Ternaryconnect,jin2018sparse} \\
Quaternary & 2   & $O(n)$      & $O(n)$      & Normal                      & Quaternary&  \cite{cheng2019uL2Q,VecQ}  \\
FixedQ     & 2-8 & $O(n)$      & $O(n)$      & Uniform                     & Fixed-point & \cite{gysel2016hardware,chen2019tdla}   \\
ZoomQ      & {2-8} & {$O(n)$}    & {$O(n)$}      & Uniform             & {\gccpf{Integer}}    & {\cite{zhou2016dorefa}}     \\
& &&& Normal& &  \\
{ClipQ}      & {2-8} & {$O(n)$}      & {$O(n)$}& Normal & {\gccpf{Integer}}     & \cite{PACT,QIL}     \\
& &&& Logistic& & \cite{wang2019haq,TSQ2018}  \\
& &&& Exponential& & \cite{cheng2019uL2Q} \\
PotQ       & 2-4 & $O(n)$      & $O(n)$      & Log-Normal                  & \gccpf{PoT} & \cite{INQ2017,ENN2017,LogQ}  \\
ResQ       & 2-8 & $O(nb)$ & $O(nb)$ & Uniform                     & \gccpf{Sum of binaries}                       & \cite{ghasemzadeh2018rebnet,LQ-Nets,ABC-Net} \\\Xhline{1.5pt}
\end{tabular}\\\flushleft\vspace{-8pt}
\gccpf{Note:}~{TC is the time complexity of quantizing scheme. $n$ is the scale of input and $b$ is the number of quantizing bits.}
{SC is the space complexity of quantizing scheme.}
{Distributions denote the distributions wanted by quantizing schemes as shown in \figref{fig:Qloss_vs_schemes_vs_bits}.}
\vspace{-5pt}
\end{table*}
\noindent\textnbf{PotQ}. \gccpf{PotQ is a non-uniform quantizing scheme that quantizes values into the form of PoT.} We define PotQ as follows:
\begin{equation}
\small
    \begin{split}
    \gccpf{\boldsymbol{d_\text{q}}}&=\alpha\cdot\text{sign}(\gccpf{\boldsymbol{d_\text{f}}})\cdot 2^e\\
    &\text{s.t.}~e=v-\mathbb{I}_{v=0},\\
    &~~~~~v=\clip{\round{\log_2(\frac{\abs(\gccpf{\boldsymbol{d_\text{f}}})}{\alpha})}}{0}{2^{b-1}-1}.
    \end{split}\nonumber
\end{equation}
Here $\mathbb{I}_{v=0}$ maps all the zeros in $v$ into ones and the non-zero values into zeros.
The quantized values and the quantizing process of PotQ are shown in \gccpf{\figref{fig:QPL_on_cifar10}(c)}.
Assuming that the inputs are normally distributed, 
we can obtain the optimal solutions offline that are $\alpha\in\{1.2240,$ $0.5181,$ $0.0381\}$ for $b\in \{2,$ $3,$ $4\}$
\footnote{\gccpf{Since the intervals of quantized values in PotQ grow exponentially  with the bitwidth increasing, the maximum bitwidth should be less than or equal to $4$~\cite{APoT}}.}. 

The quantization loss comparisons of the above candidate schemes handling different distributions are shown in \figref{fig:Qloss_vs_schemes_vs_bits}, and the detailed comparisons of these schemes are presented in \tabref{tab:candidate_scheme_table}.
The results show that the schemes perform widely divergent on different distributions.
No scheme can always achieve the minimum quantization loss across various distributions. 
However, the weights and activation of each layer in DNNs tend to distribute differently. 
This observation implies that employing an undesirable quantizing scheme will make it difficult to fit the distributions and degrade the accuracy of DNNs.
Therefore, seeking desirable quantization for specific layers or models is strongly demanded.

\begin{figure}[H]
    \centering
    \includegraphics[width=1\columnwidth]{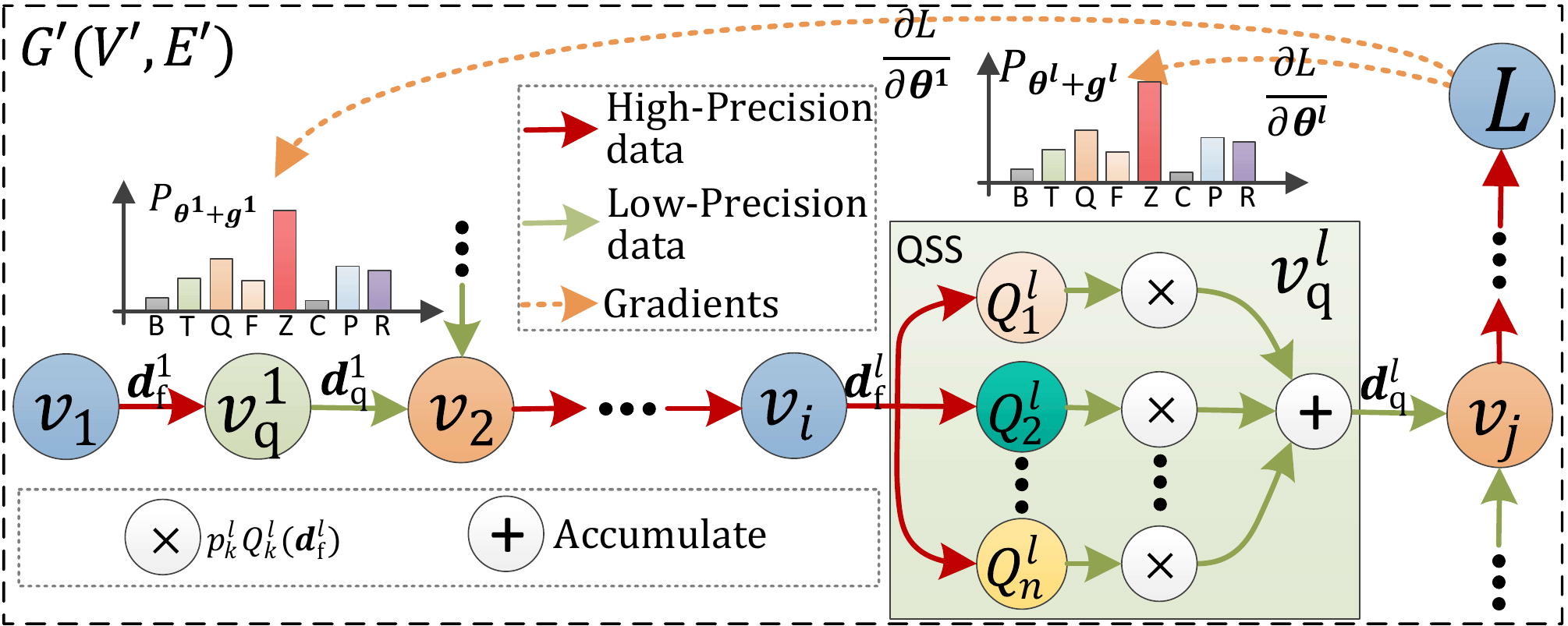}
    \caption{The quantizing scheme search process. Reparameterization is adopted and we optimize the state parameters to seek desirable quantizing schemes.}
    \label{fig:automatic_scheme_search}
\end{figure}
\subsection{Seeking Scheme}
We denote the computing graph of a neural architecture as $G(V,E)$ and the corresponding quantized graph as $G'(V',E')$. 
$G'$ is generated from $G$ by adding the quantizing vertices, that is, $V\subset V'$. $V'-V$ is the set that consists of the quantizing vertices. $N=|V'-V|$ is the number of the added quantizing vertices.
We seek suitable schemes for these quantizing vertices to reduce the accuracy degradation in quantization.

We denote all the schemes introduced in \subsecref{sec:candidate_schemes} as a candidate set $\{Q_1^l,$ $Q_2^l,$ $Q_3^l,$ $\cdots,$ $Q_n^l\}$.
Let $v_i, v_j\in V'$, $v_q^l\in V'-V$ be the $l$-th quantizing vertex, and $\edge{v_i,v_q^l},\edge{v_q^l,v_j}\in E'$. 
The implementation of $v_q^l$ should be a scheme $Q_k^l$ selected from the candidate set.
Let $\gccpf{\boldsymbol{d_\text{f}}^l}$ and $\gccpf{\boldsymbol{d_\text{q}}^l}$ denote the original vector and quantized vector, respectively.
A quantizing process of the scheme $Q_k^l$ is denoted as $\boldsymbol{d_\text{q}}^l = Q_k^l(\gccpf{\boldsymbol{d_\text{f}}^l})$. 

Quantizing scheme search (QSS) is proposed to seek a desirable quantizing scheme $Q_k^l$ from the candidate set and quantize $\gccpf{\boldsymbol{d_\text{f}}^l}$ into $\gccpf{\boldsymbol{d_\text{q}}^l}$ with small accuracy degradation.
We employ the sampling way as described in DNAS~\cite{wu2018mixed} to find desirable schemes.
As shown in \figref{fig:automatic_scheme_search}, we construct state parameters $\boldsymbol\theta^l=[\theta_1^l,$ $\theta_2^l,$ $\cdots,$ $\theta_n^l]^T$ that correspond to the quantizing schemes, and sample a quantizing scheme with the probabilities $P_{\boldsymbol\theta}$ before each training phase.
The sampling process can be defined as follows:
\begin{equation}
    \begin{split}
        \gccpf{\boldsymbol{d_\text{q}}^l}&=Q_k^l(\gccpf{\boldsymbol{d_\text{f}}^l})\\
        &\text{s.t.}~k\sim P_{\boldsymbol\theta^l},~P_{\boldsymbol\theta^l}=\text{softmax}(\boldsymbol\theta^l).
    \end{split}\nonumber
\end{equation}
Here $\text{softmax}(\cdot)$ maps values into probabilities.

Unfortunately, the conventional sampling process $k\sim P_{\boldsymbol\theta^l}$ is non-differentiable. 
It means that the sampling result is hard to guide the optimization of state parameters $\boldsymbol\theta^l$.
To address this issue, we employ the Gumbel-Softmax~\cite{gumbel_softmax1,gumbel_softmax2} to realize a differentiable sampling process as follows:
\begin{equation}
\label{eq:gumble_sampling}
\begin{split}
    \gccpf{\boldsymbol{d_\text{q}}^l}=&Q_k^l(\gccpf{\boldsymbol{d_\text{f}}^l})\\
    \text{s.t.}&~k=\underset{k}{\arg}\max(P_{\boldsymbol\theta^l+\boldsymbol{g}^l}),\\
    &P_{\boldsymbol\theta^l+\boldsymbol{g}^l}=\text{softmax}(\boldsymbol\theta^l+\boldsymbol{g}^l),~\boldsymbol{g}^l\sim \text{Gumbel}(0,1)^n.
\end{split}
\end{equation}
Here $g_k^l\in \boldsymbol{g}^l$ is a value drawn from the Gumbel distribution. 
The sampling process in \eqref{eq:gumble_sampling} is differentiable and the gradient of $\gccpf{\boldsymbol{d_\text{q}}^l}$ with respect to $\theta_k^l\in\boldsymbol\theta^l$ can be derived.
However, the gradient of the "hard" sampling process of $\arg\max(\cdot)$ is zero everywhere, and the state parameters can not be optimized.
To solve this problem, we introduce a "soft" sampling process that adopts the weighted sum of $p_k^l$ and $Q_k^l(\gccpf{\boldsymbol{d_\text{f}}^l})$. Finally, the seeking scheme of QSS is defined as follows:
\begin{equation}
    \label{eq:gumble_weight_sum}
    \begin{split}
    \gccpf{\boldsymbol{d_\text{q}}^l}&=\sum_{k=1}^{n}p_k^l\cdot Q_k^l(\gccpf{\boldsymbol{d_\text{f}}^l})\\
    &\text{s.t.}~p_k^l\in\text{softmax}(\frac{\boldsymbol\theta^l+\boldsymbol{g}^l}{\tau}),
    ~\boldsymbol{g}^l\sim \text{Gumbel}(0,1)^n.
    \end{split}
\end{equation}
Here $\tau$ is a temperature coefficient. As $\tau\to \infty$, each output of $Q_k^l(\gccpf{\boldsymbol{d_\text{f}}^l})$ has the same weight $p_k^l$ and the $\gccpf{\boldsymbol{d_\text{q}}^l}$ equals the averaged outputs. Thus all of the state parameters can be optimized simultaneously. As $\tau\to 0$, the $\gccpf{\boldsymbol{d_\text{q}}^l}$ is equivalent to the sampling result in \eqref{eq:gumble_sampling}.
We smoothly decay $\tau$ from a large value to 0 to select a desirable quantizing scheme for $v_q^l$. The decaying strategy of $\tau$ is defined as follows:
\begin{equation}
\label{eq:temp_decay}
    \tau=\tau_0\cdot(1 - \delta /\Delta)^{p}.
\end{equation}
$\tau_0$ is the initial temperature. $\delta$ is the current training epoch and $\Delta$ is the number of total training epochs. $p$ is the exponential coefficient.

Let the final loss of $G'(V',E')$ be $L$.
Since \eqref{eq:gumble_weight_sum} is differentiable, the gradients of $L$ with respect to $\boldsymbol{\theta}^l$ can be computed as follows:
\begin{equation}
    \frac{\partial L}{\partial\boldsymbol{\theta}^l}=
    \frac{\partial L}{\partial \gccpf{\boldsymbol{d_\text{q}}^l}}\frac{\partial \gccpf{\boldsymbol{d_\text{q}}^l}}{\partial\boldsymbol{\theta}^l}. \nonumber
\end{equation}
It means that a gradient-descent algorithm can be employed to optimize the state parameters $\boldsymbol{\theta}^l$.
Therefore, we can seek a quantizing scheme for each quantizing vertex $v_q^l$ to minimize the loss of neural architecture, so as to improve accuracy.

In addition, different quantizing schemes require distinct accelerating implementations, such as the bit-operations for binaries/ternaries, low-bit multiplication for fixed-point values, and shift operations for PoTs. Implementing all of these operations is unreasonable for resource-constrained devices. 
To simplify the hardware implementation of accelerators, searching for one shared quantizing scheme for all the vertices in one neural architecture is still demanded.
We realize the coarse-grained QSS by maintaining and sharing one group of state parameters $\boldsymbol\theta$ for all quantizing vertices in $G'(V',E')$.
In coarse-grained QSS, 
the gradients of $L$ with respect to $\boldsymbol\theta$ is computed as follows:
\begin{equation}
    \frac{\partial L}{\partial\boldsymbol{\theta}}=
    \sum_{l=1}^{N}
    \frac{\partial L}{\partial \gccpf{\boldsymbol{d_\text{q}}^l}}\frac{\partial \gccpf{\boldsymbol{d_\text{q}}^l}}{\partial\boldsymbol{\theta}}
    . \nonumber
\end{equation}
\begin{figure}[H]
    \centering
    \includegraphics[width=0.8\columnwidth]{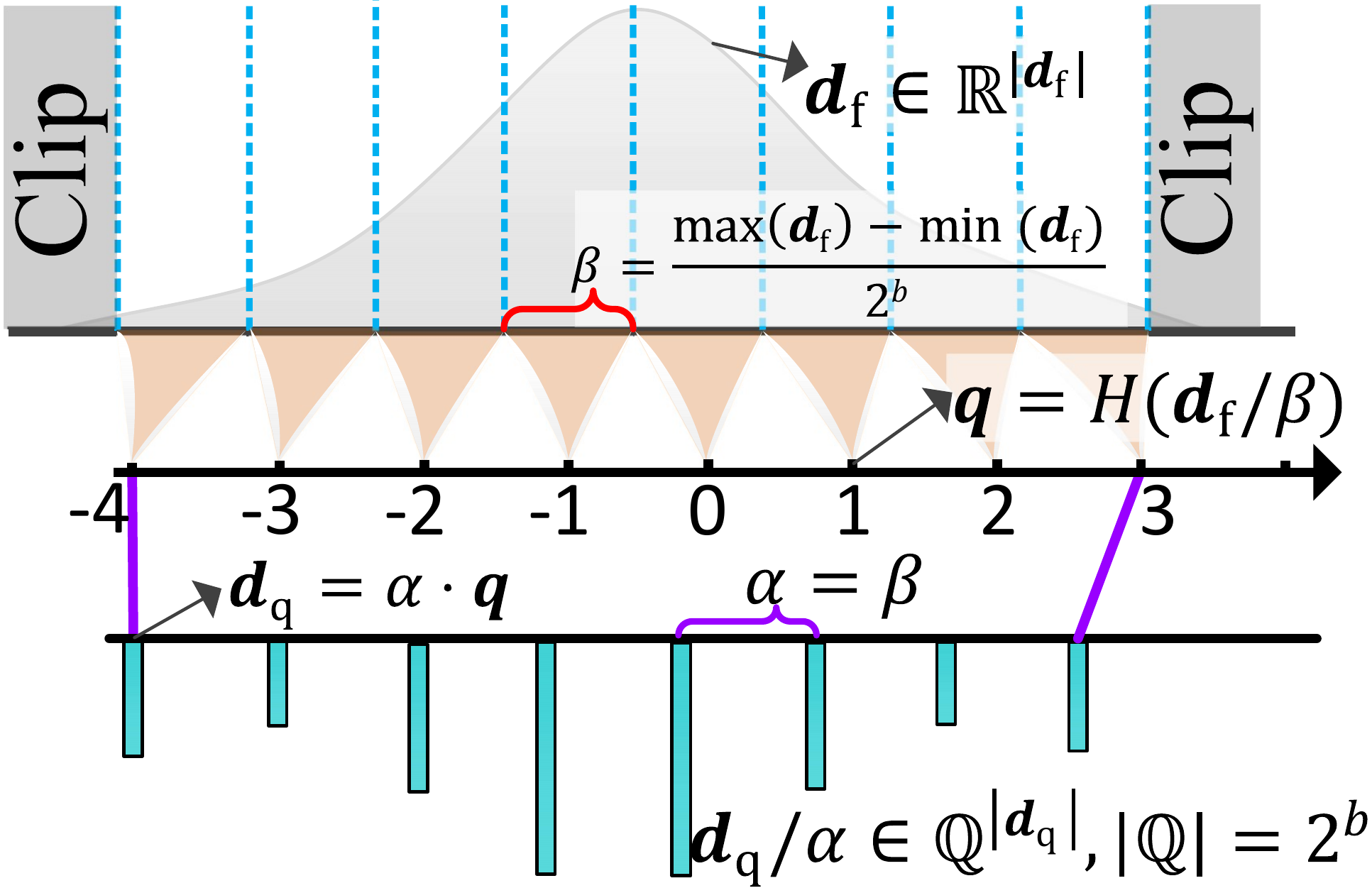}
    \caption{\label{fig:automatic_precision_search}A general quantizing process: mapping full-precision values to integers and then scaling these integers into a target range.}
\end{figure}
\section{Quantizing Precision Learning}\label{sec:qpl}
Quantizing precision, i.e., the bitwidth, is an essential attribute of quantizing schemes and determines the number of quantized values. 
Selecting an optimal precision for a specific quantizing scheme is vital for balancing the efficiency and accuracy of quantized DNNs.
In this section, we reparameterize the quantizing precision and learn the relatively optimal mixed-precision model within limited model size and memory footprint.

\subsection{Bitwidth Reparameterization}
Let $b$ be the number of the quantizing bits and $\mathbb{Q}$ be the quantized value set, and we have $|\mathbb{Q}|=2^b$. 
A general quantizing process of $\gccpf{\boldsymbol{d_\text{q}}}=Q_k(\gccpf{\boldsymbol{d_\text{f}}})$ can be represented as two stages: 1) mapping the full-precision weights or activation to the quantized values that belong to $\mathbb{Q}$, and then 2) scaling the quantized values into a target range, as shown in \figref{fig:automatic_precision_search}.
Here we expand the general quantizing process as following formulations.
\begin{equation}
\label{eq:projection_scheme}
    \begin{split}
    \gccpf{\boldsymbol{d_\text{q}}}&=Q_k(\gccpf{\boldsymbol{d_\text{f}}})=\alpha H(\frac{\gccpf{\boldsymbol{d_\text{f}}}}{\beta})\\
        &\text{s.t.}~H:\mathbb{R}^{|\gccpf{\boldsymbol{d_\text{f}}}|}\rightarrow \mathbb{Q}^{|\gccpf{\boldsymbol{d_\text{q}}}|},~|\mathbb{Q}|=2^b.
    \end{split}
\end{equation}
Here $|\gccpf{\boldsymbol{d_\text{f}}}|$ and $|\gccpf{\boldsymbol{d_\text{q}}}|$ are the number of elements of $\gccpf{\boldsymbol{d_\text{f}}}$ and $\gccpf{\boldsymbol{d_\text{q}}}$, respectively, and $|\gccpf{\boldsymbol{d_\text{f}}}|=|\gccpf{\boldsymbol{d_\text{q}}}|$. $\mathbb{R}$ is the set of real numbers. 
$H$ is a function that projects the vector with real values into that with quantized values. 
$\alpha$ is the scaling factor and $\beta$ is the average step size of the quantizing scheme $Q_k$. 
In general, $\alpha$ varies linearly with $\beta$ to maintain a similar range of $\gccpf{\boldsymbol{d_\text{f}}}$ and $\gccpf{\boldsymbol{d_\text{q}}}$, thereby reducing quantization loss.
\begin{equation}
    \alpha=\lambda\beta. \nonumber
\end{equation}
Typically, $\lambda=1$ and $\alpha=\beta$~\cite{zhou2016dorefa,QAT,wang2019haq,cheng2019uL2Q}. 
The average step size $\beta$ can be calculated using the quantizing precision $b$ and the range of $\gccpf{\boldsymbol{d_\text{f}}}$ as follows:
\begin{equation}
    \beta=\frac{\max(\gccpf{\boldsymbol{d_\text{f}}})-\min(\gccpf{\boldsymbol{d_\text{f}}})}{2^b}.\nonumber
\end{equation}
Therefore, the loss $L$ of $G'(V',E')$ is differentiable with respect to the quantizing precision $b$. The gradient $\frac{\partial L}{\partial b}$ can be calculated as follows:
\begin{equation}
\label{eq:differential_precision}
\small
    \begin{split}
    \frac{\partial L}{\partial b}&=
    \frac{\partial L}{\partial \gccpf{\boldsymbol{d_\text{q}}}}\frac{\partial \gccpf{\boldsymbol{d_\text{q}}}}{\partial \alpha}\frac{\partial\alpha}{\partial\beta}\frac{\partial \beta}{\partial b}\\
    &=-\frac{\partial L}{\partial \gccpf{\boldsymbol{d_\text{q}}}}\frac{\max(\gccpf{\boldsymbol{d_\text{f}}})-\min(\gccpf{\boldsymbol{d_\text{f}}})}{2^{b}(\ln{2})^{-1}}\cdot(H(\frac{\gccpf{\boldsymbol{d_\text{f}}}}{\lambda\alpha})-\frac{\gccpf{\boldsymbol{d_\text{f}}}}{\lambda\alpha}).
    \end{split}
\end{equation}
We employ the straight through estimator (STE) ~\cite{zhou2016dorefa} to compute the differential of $H(\cdot)$, i.e., $H'(\cdot)=1$.
Based on \eqref{eq:differential_precision}, we get the gradient of $L$ with respect to the quantizing precision $b$, which means that we can reparameterize $b$ and employ the gradient-descent algorithm to learn a reasonable quantizing precision.  

\subsection{Precision Loss}
Generally, DNNs tend to learn high-precision weights or activation by minimizing $L$. 
It implies that the values of the quantizing precision $b$ for DNNs will be optimized to be great ones, such as 32 bits or 64 bits, instead of small ones.
To learn a policy with low average quantizing precision,
we propose precision loss $\overline{L}$, which measures the distance between the average quantizing precision and a target precision $\overline{b}$.
\begin{equation}
\label{eq:average_bits}
\begin{split}
    \overline{L}&=(E(\mathbb{B})-\overline{b})^2\\
    &\text{s.t.}~\mathbb{B}=\{b_i|i=1,2,\cdots,\sum_{l=1}^N|\gccpf{\boldsymbol{d_\text{q}}}(l)|\}.
\end{split}
\end{equation}
$\overline{b}$ is the expected quantizing bitwidth of a neural architecture, and is preset before a precision learning phase.
$b_i$ is the bitwidth of one weight or activation value in the neural architecture $G'(V',E')$. $\mathbb{B}$ is the set that consists of $b_i$. $N=|V'-V|$ is the number of quantizing vertices as described in \secref{sec:qss}.
The gradient of $L+\overline{L}$ with respect to $b$ is defined as follows:
\begin{equation}
    g_{b}=\frac{\partial L}{\partial b}+\frac{\partial \overline{L}}{\partial b}.\nonumber
\end{equation}
\section{Implementation}\label{sec:qag}
We denote the computing graph of a neural architecture as $G(V,E)$. $V$ is the set of the vertices and $v_i\in V$ indicates a vertex. 
$E$ is the set of edges and $\edge{v_i,v_j}\in E$ indicates that there is a tensor $\gccpf{\boldsymbol{d}^{i,j}}$ flowing from $v_i$ to $v_j$.  
$\text{Id}(v_i)$ and $\text{Od}(v_i)$ compute the in-degree and out-degree of $v_i$, respectively. 
$v_i$ is a data vertex when $\text{Id}(v_i)=0$, such as input image, feature map, and weight. 
$v_i$ is an operation vertex when $\text{Id}(v_i)>0$, such as Convolution, Batch Normalization, and Full-Connection, and so on.

\renewcommand{\arraystretch}{1.0}
\begin{algorithm}[H] 
\caption{\label{alg:model_generation}Quantized Architecture Generation (QAG).}
\begin{algorithmic}[1] 
\REQUIRE Computing graph $G(V,E)$, expensive vertex set $\gccpf{V_\text{e}} \subset V$.
\ENSURE Quantized architecture $G'(V',E')$.
\STATE $V'\leftarrow\{v_i|v_i\in V, Id(v_i)=0\}$.
\STATE $E'\leftarrow\{\}$
\STATE $I\leftarrow\{v_j|v_i\in V',v_j\in V-(V'\cap V), \edge{v_i,v_j}\in E\}$
\WHILE{$I \neq \phi$}
    \FOR{$v_j\in I$}
        \STATE $I'\leftarrow\{v_k|v_k\in V, \edge{v_k,v_j}\in E\}$
        \IF{$I'\subseteq V'$}
            \STATE $V'\leftarrow V' \cup \{v_j\}$
            \FOR{$v_k \in I'$}
                \IF{$v_j \in \gccpf{V_\text{e}}$}
                    \STATE create quantizing vertex $v_{\text{q}_k}$
                    \STATE $V'\leftarrow V'\cup \{v_{\text{q}_k}\}$
                    \STATE $E'\leftarrow E'\cup \{\edge{v_k,v_{\text{q}_k}},\edge{v_{\text{q}_k},v_j}\}$
                \ELSE 
                    \STATE $E'\leftarrow E'\cup \{\edge{v_k,v_j}\}$
                \ENDIF
            \ENDFOR
        \ENDIF
    \ENDFOR
    \STATE $I\leftarrow\{v_j|v_i\in V',v_j\in V-(V'\cap V), \edge{v_i,v_j}\in E\}$
\ENDWHILE
\end{algorithmic} 
\end{algorithm}

\subsection{Quantized Architecture Generation}
Let $\gccpf{\boldsymbol{d}^{i,j}}$ denote a tensor that flows into a time-consuming and/or memory-consuming vertex, such as the weight tensor of convolutional (Conv) vertices and full-connected (FC) vertices.
The target of quantization is to reduce the precision of $\gccpf{\boldsymbol{d}^{i,j}}$ and shrink the computing consumption of DNNs.
These time-consuming and/or memory-consuming vertices are called expensive vertices.
$\gccpf{V_\text{e}}\subset V$ is the expensive vertex set to be quantized. 
The Conv and FC vertices occupy more than 99\% of operations and memory footprint in mainstream DNN architectures and applications.
According to the above observations, we set the Conv and FC vertices as the default expensive vertices,
that is, $\gccpf{V_\text{e}}=\{v_i| v_i \in V, v_i~\text{is Conv or FC}\}$.
\begin{figure*}
    \centering
    \includegraphics[width=0.85\textwidth]{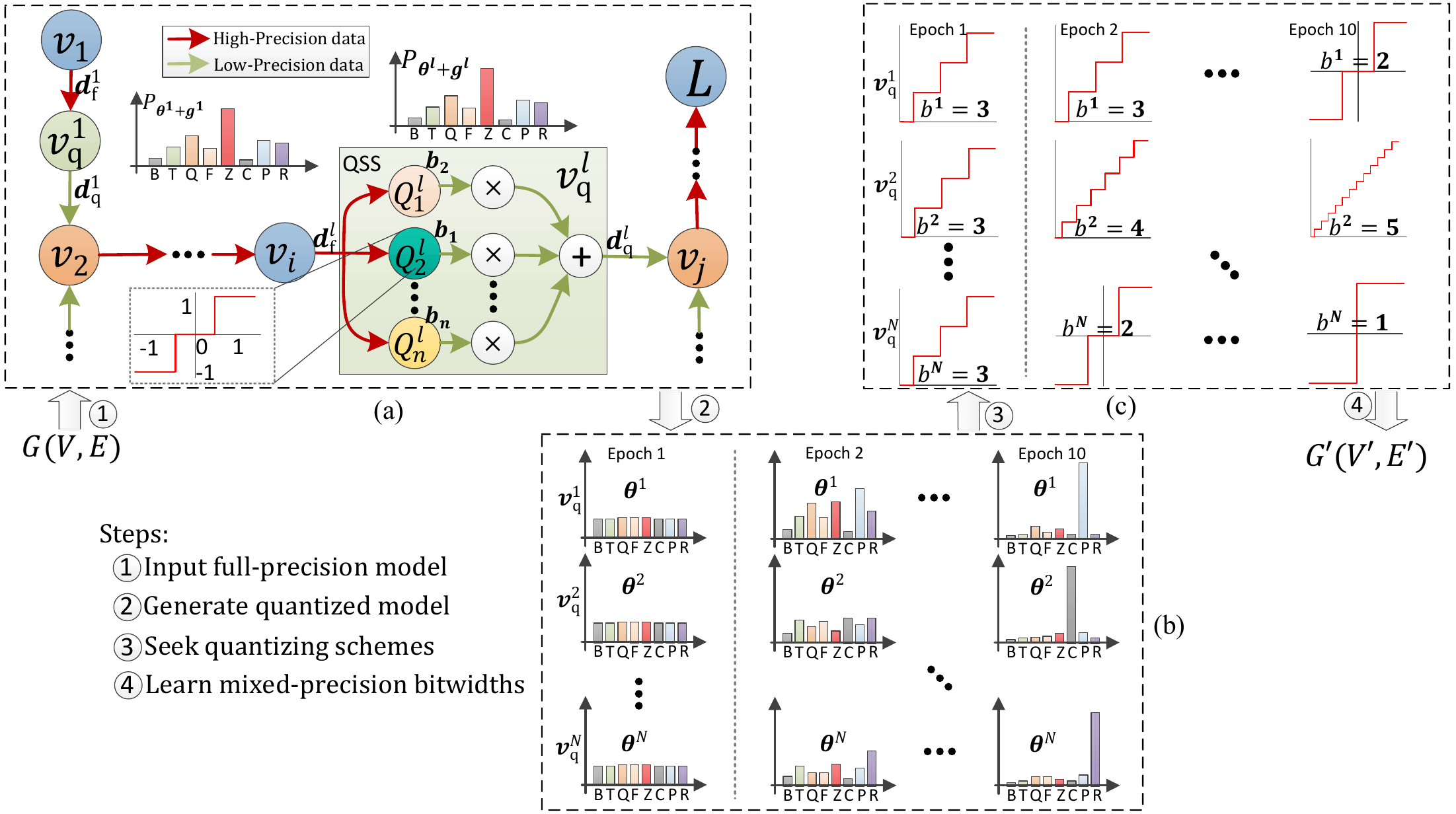}
    \caption{End-to-end neural network quantization. Input is a full-precision architecture $G(V,E)$ and output is the efficient quantized architecture $G'(V',E')$ with desirable quantizing strategies. \gccpf{(a) Generating a quantized architecture with QAG. (b) Seeking quantizing schemes with QSS. (c) Learning precision with QPL.}}
    \label{fig:autoqnn}
\end{figure*}

We design the quantized architecture generation (QAG) algorithm to automatically reconstruct the computing graph $G(V,E)$ into its quantized counterpart $G'(V',E')$ that requires less memory footprint and few computing overheads.
As shown in \algref{alg:model_generation}, 
for a specific DNN with computing graph $G(V,E)$ and the expensive vertex set $\gccpf{V_\text{e}}$, QAG first collects the vertices with 0 in-degree as the initial vertex set $V'$ and initializes the edge set $E'=\{\}$. 
Let $I=\{v_j|v_i\in V', v_j\in V-(V'\cap V), \edge{v_i,v_j}\in E\}$, we have $I\neq \phi$ when $V \neq V'\cap V$, since $G=(V,E)$ is a connected graph.
Then we move the vertices from $I$ to $V'$ iteratively.
During this process, we embed quantizing vertices into the edges flowing into expensive vertices.

The original $G(V,E)$ can be automatically reconstructed as $G'(V',E')$ within $|V|$ iterations, and the tensors flowing into the expensive vertices of $V_\text{e}$ are replaced with low-precision representations by embedding quantizing vertices. 
Therefore, the new architecture $G'(V',E')$ requires less computing budget than $G(V,E)$.
The average time and space complexity of \algref{alg:model_generation} are $O(|V||E|)$ and $O(|V|+|E|)$, respectively.

\subsection{End-to-End Framework}
The proposed QAG in \algref{alg:model_generation} can significantly reduce the quantization workload and avoid prone errors, so as to provide fast, cheap, and reliable quantized architecture generation. We finally implement the end-to-end framework based on QAG, QSS, and QPL.
As shown in \figref{fig:autoqnn} and \algref{alg:model_training}, AutoQNN is constructed with three stages and takes a full-precision architecture as input. 
AutoQNN first reconstructs the input architecture into a quantized one through QAG.
Then it trains total $\Delta$ epochs to seek desirable quantizing schemes through QSS automatically. 
Finally, AutoQNN retrains/fine-tunes the quantized architecture with QPL to converge. 
\renewcommand{\arraystretch}{1.0}
\begin{algorithm}[H] 
\caption{\label{alg:model_training}AutoQNN Framework.}
\begin{algorithmic}[1] 
\REQUIRE Computing architecture $G(V,E)$, expensive vertices $\gccpf{V_\text{e}}\subset V$, candidate quantizing schemes $Q$, Dataset $\mathcal{X}$, training epochs $\Delta$.
\ENSURE Quantized architecture $G'(V',E')$.
\STATE Generating $G'(V',E')$ with \algref{alg:model_generation}.
\FOR{$\delta=1$ to $\Delta$}
\STATE Decaying $\tau$ with ~\eqref{eq:temp_decay}.
\STATE Training $G'$ on $\mathcal{X}$ with respect to weights and $\boldsymbol{\theta}$.
\ENDFOR
\STATE Sampling the quantizing schemes.
\STATE Training the weights and $b$ of $G'$ on $\mathcal{X}$ to converge.
\end{algorithmic} 
\end{algorithm}

\section{Evaluation}\label{sec:exp}
In order to train DNN models and evaluate their performance conveniently, we implement AutoQNN
and integrate it into Keras\footnote{https://github.com/fchollet/keras. Oct. 2022.\label{keras}} (v2.2.4). 
Furthermore, we \gccpf{implement} the eight candidate quantizing schemes presented in \secref{sec:qss}. 
For ease of notation, we use the combination of scheme name and bitwidth to denote one quantizing strategy. 
For example, "P-3" indicates a PotQ quantizer with a bitwidth of 3. 
All quantized architectures are automatically constructed by AutoQNN in our experiments.
In addition, respecting that most DNNs have employed ReLU~\cite{nair2010rectified} to eliminate the negative elements of activation, we do not use binary, ternary, and quaternary schemes in activation quantization. 

\begin{table*}[!t]
\centering
\caption{\gccpf{Optimal Quantizing Strategies for AlexNet and ResNet18}\label{tab:best_quantizing_scheme_precison}}
\setlength{\tabcolsep}{15pt}
\begin{tabular}{lcccc}\Xhline{1.5pt}
{Models} & \gccpf{Notation} & Quantizing Bits of & Quantizing Bits of & Average\\
 & \gccpf{(W/A)} &  Layer Weights & Layer Activation & \gccpf{Bits (W/A)} \\\Xhline{.5pt}
{AlexNet} & P-2/C-2 & \gccpf{4, 4, 4, 4, 2, 2} & \gccpf{2, 2, 2, 2, 3, 3} & 2.13/2.01 \\
 & P-3/C-3 & \gccpf{4, 4, 4, 4, 3, 3} & \gccpf{3, 3, 4, 3, 4, 4} & 3.06/3.20 \\
 & P-4/C-4 & \gccpf{4, 4, 4, 4, 4, 4} & \gccpf{3, 3, 3, 5, 5, 5} & 4.00/4.05 \\
{ResNet18} & {P-2/C-2} & \gccpf{4, 3, 4, 4, 3, 3, 4, 4, 2, 2,} & \gccpf{3, 2, 3, 2, 2, 3, 3, 2, 2, 2,} & {2.13/2.49} \\
& &\gccpf{4, 2, 2, 2, 2, 2, 2, 4, 2, 2} & \gccpf{3, 3, 2, 3, 3, 3, 3, 3, 3, 3} & \\
 & {P-3/C-3} & \gccpf{4, 4, 4, 4, 4, 4, 4, 4, 4, 4,} & \gccpf{4, 3, 3, 2, 2, 4, 4, 2, 2, 3,} & {2.94/3.07} \\
& & \gccpf{4, 4, 4, 4, 4, 4, 2, 4, 3, 2} &\gccpf{4, 4, 2, 3, 4, 5, 6, 4, 5, 4} & \\
 & {P-4/C-4} & \gccpf{4, 4, 4, 4, 4, 4, 4, 4, 4, 4,} & \gccpf{5, 4, 5, 3, 3, 4, 4, 3, 3, 3,} & {4.00/4.08} \\
 & &\gccpf{4, 4, 4, 4, 4, 4, 4, 4, 4, 4} &\gccpf{5, 6, 3, 4, 5, 7, 7, 5, 7, 6} & \\\Xhline{1.5pt}
\end{tabular}\\\flushleft\vspace{-8pt}
\gccpf{Note:~{The first and last layers are not quantized. W/A denotes the results for Weights and Activation, respectively.}}
{The maximum quantizing bitwidth of PotQ is 4.}
\end{table*}
\subsection{Image Classification}
Image classification provides the basic cues for many computer vision tasks, so the results of image classification are representative for the evaluation of quantization.
For a fair comparison with the~\sArt~\gccpf{quantization}~\cite{APoT,QIL}, we quantize the weights and activation of all layers except the first and last layers. 

\subsubsection{Dataset and Models}
We conduct experiments with widely used models on ImageNet~\cite{deng2009imagenet} dataset, including AlexNet~\cite{alexnet}, ResNet18~\cite{he2016resnet}, ResNet50~\cite{he2016resnet}, MobileNets~\cite{mobilenet}, MobileNetV2~\cite{sandler2018mobilenetv2}, and InceptionV3~\cite{inceptionv3}. 
\gcc{To make a fair comparison and ensure reproducibility, the full-precision models used in our experiments refer to the full-precision pre-trained models obtained from open sources.
Specifically, AlexNet refers to~\cite{simon2016cnnmodels}, and ResNet18 cites from
literature~\cite{gross2016training}. 
Other models are obtained from official Keras community\textsuperscript{\ref{keras}}.
Referenced accuracy results of the full-precision models used in our experiments are also from the open-source results, and no further fine-tuning is made.}
The data argumentation of ImageNet can be found here\footnote{https://github.com/tensorflow/models. Oct. 2022.}.

\subsubsection{Comparison With~State-of-the-Art~\gccpf{Work}}\label{sec:comparing_with_fixed_prcision}

\cumparagraph{Quantizing Strategies}

For quantizing scheme search, we configure all the candidate schemes with the same \gcc{low} bitwidth of 3\gcc{, since low-bitwidth quantization can highlight the difference among candidate schemes,}\gcc{ thus ensuring the robustness of searching results}.
We train the models with coarse-grained QSS by 10 epochs to find desirable quantizing schemes. 
Then, based on the found quantizing schemes, we train 60 epochs to learn a reasonable quantizing precision for each layer.

The quantizing strategies found by QSS and QPL are shown in \tabref{tab:best_quantizing_scheme_precison}.
On both AlexNet and ResNet18, PotQ is the desirable scheme for weight quantization, and ClipQ is the scheme for activation quantization.
The reason for the results is that most of the weights are Log-Normal distributed, and PotQ is the best one for handling this distribution, as described in \secref{sec:qss}.
The activation distributions of ResNet18 and AlexNet are mostly bell-shaped, and ClipQ performs well on bell-shaped distributions, such as Normal, Logistic, and Exponential distributions described in \secref{sec:qss}. 
Therefore, PotQ and ClipQ are selected as the solutions for the weight and activation quantization, respectively.

Besides, QPL has learned mixed-precision strategies.
For a fair comparison with the \sArt~fixed-precision schemes, we compute the average bitwidth for representing weights and activation of models as conditions in our experiments, since the models with the same average bitwidth consume the same memory footprint and storage in model inference.

\cumparagraph{Comparison Results} 

We compare AutoQNN with the \sArt~fixed-precision \gccpf{quantization} across various quantizing bits and the results are shown in \tabref{tab:compare_with_sArts}.
The results show that AutoQNN can consistently outperform the \sArt~\gccpf{methods} with much higher inference accuracies.
At extreme conditions with only 2 bits for weights and activation, the accuracy of AutoQNN on AlexNet is \text{59.75\%}, which is slightly lower than the accuracy of the \gcc{referenced} full-precision AlexNet by \text{0.26\%}. 
The accuracy also exceeds that of QIL by \text{1.65\%}, and outperforms that of VecQ by \text{1.27\%}.
Besides, AutoQNN achieves an accuracy of \text{68.84\%} on ResNet18 using only 2 bits for all weights and activation. 
The result is slightly less than the accuracy of the \gcc{referenced} full-precision model by \text{0.76\%} and exceeds APoT and VecQ by \text{1.74\%} and \text{0.61\%}, respectively.
AutoQNN achieves the highest accuracy results of \text{61.85\%} and \text{62.63\%} when quantizing AlexNet into a 3-bit model and a 4-bit model correspondingly.
AutoQNN also realizes the best results of \text{69.88\%} and \text{70.36\%} when quantizing the weights and activation of ResNet18 into 3 bits and 4 bits correspondingly. 
\gcc{It is worth noting that the results of quantized models may exceed the specific referenced accuracy results by fine-tuning. Still, quantization can harm accuracy, and the quantized models cannot perform better than full-precision models in accuracy theoretically.}

The experiments demonstrate that AutoQNN can automatically seek desirable quantizing strategies to reduce accuracy degradation under the different conditions of model size and memory footprint, thus achieving a new balance between accuracy and efficiency in DNN quantization.

\renewcommand{\arraystretch}{1.1}
\begin{table}[H]
\centering
\caption{\gccpf{Accuracy Comparison With State-of-the-Art~Methods}\label{tab:compare_with_sArts}}
\setlength{\tabcolsep}{2.5pt}
\begin{tabular}{lcccccc}\Xhline{1.5pt}
QB& Methods & {W/A} & \multicolumn{2}{c}{AlexNet} & \multicolumn{2}{c}{ResNet18}\\\Xhline{.5pt}
32&\gcc{Referenced} & 32/32 & 60.01 & 81.90 & 69.60 & 89.24 \\
2&TWNs~\cite{TWNs} & 2/32 & 57.50 & 79.80 & 61.80 & 84.20 \\
&TTQ~\cite{zhu2016TTQ} & 2/32 & 57.50 & 79.70 & 66.60 & 87.20 \\
&INQ~\cite{INQ2017} & 2/32 & - & - & 66.02 & 87.12 \\
&ENN~\cite{ENN2017} & 2/32 & 58.20 & 80.60 & 67.00 & 87.50 \\
&uL2Q~\cite{cheng2019uL2Q} & 2/32 & - & - & 65.60 & 86.12 \\
&VecQ~\cite{VecQ} & 2/32 & 58.48 & 80.55 & 68.23 & 88.10 \\
&TSQ~\cite{TSQ2018} & 2/2 & 58.00 & 80.50 & - & - \\
&DSQ~\cite{DSQ} & 2/2 & - & - & 65.17 & - \\
&PACT~\cite{PACT} & 2/2 & 55.00 & 77.70 & 64.40 & 85.60 \\
&Dorefa-Net~\cite{zhou2016dorefa} & 2/2 & 46.40 & 76.80 & 62.60 & 84.40 \\
&LQ-Nets~\cite{LQ-Nets} & 2/2 & 57.40 & 80.10 & 64.90 & 85.90 \\
&QIL~\cite{QIL} & 2/2 & 58.10 & - & 65.70 & - \\
&APoT~\cite{APoT} & 2/2 & - & - & 67.10 & 87.20 \\
&\gcc{BRQ~\cite{residualquantization2021tom}} & \gcc{2/2} & - & - & \gcc{64.40} & - \\
&\gcc{TRQ~\cite{residualquantization2021tom}} & \gcc{2/2} & - & - & \gcc{63.00} & - \\
&\text{AutoQNN} & P-2/C-2 & \underline{59.75} & \underline{81.72} & \underline{68.84} & \underline{88.50} \\
3&INQ & 3/32 & - & - & 68.08 & 88.36 \\
&ENN-2 & 3/32 & 59.20 & 81.80 & 67.50 & 87.90 \\
&ENN-4 & 3/32 & 60.00 & 82.40 & 68.00 & 88.30 \\
&VecQ & 3/32 & 58.71 & 80.74 & 68.79 & 88.45 \\
&DSQ & 3/3 & - & - & 68.66 & - \\
&PACT & 3/3 & 55.60 & 78.00 & 68.10 & 88.20 \\
&Dorefa-Net & 3/3 & 45.00 & 77.80 & 67.50 & 87.60 \\
&ABC-Net & 3/3 & - & - & 61.00 & 83.20 \\
&LQ-Nets & 3/3 & - & - & 68.20 & 87.90 \\
&QIL & 3/3 & 61.30 & - & 69.20 & - \\
&APoT & 3/3 & - & - & 69.70 & 88.90 \\
&\gcc{BRQ} & \gcc{3/3} & - & - & \gcc{66.10} & - \\
&\text{AutoQNN} & P-3/C-3& \underline{61.85} & \underline{83.47} & \underline{69.88} & \underline{89.07} \\
4&INQ & 4/32 & - & - & 68.89 & 89.01 \\
&uL2Q & 4/32 & - & - & 65.92 & 86.72 \\
&VecQ & 4/32 & 58.89 & 80.88 & 68.96 & 88.52 \\
&DSQ & 4/4 & - & - & 69.56 & - \\
&PACT & 4/4 & 55.70 & 78.00 & 69.20 & 89.00 \\
&Dorefa-Net & 4/4 & 45.10 & 77.50 & 68.10 & 88.10 \\
&LQ-Nets & 4/4 & - & - & 69.30 & 88.80 \\
&QIL & 4/4 & 62.00 & - & 70.10 & - \\
&BCGD~\cite{BCGD} & 4/4 & - & - & 67.36 & 87.76 \\
&\gcc{TRQ~\cite{residualquantization2021tom}} & \gcc{4/4} & - & - & \gcc{65.50} & - \\
&\text{AutoQNN} & P-4/C-4 & \underline{62.63} & \underline{83.93} & \underline{70.36} & \underline{89.43} \\\Xhline{1.5pt}
\end{tabular}
\\\flushleft\vspace{-8pt}
\gccpf{Note: QB is the quantizing bitwidth. W/A denotes the weight/activation bitwidth. The best results are \underline{underlined}.}
\end{table} 
\begin{table}[H]
\caption{\label{tab:comparing_with_mixedp}\gccpf{Accuracy Comparison With Mixed-Precision Quantization}}
\setlength{\tabcolsep}{2pt}
\begin{tabular}{llcc}
\Xhline{1.5pt}
Models & Methods & W/A & \gccpf{Accuracy} \\
\Xhline{.5pt}
{ResNet18} & MixedP~\cite{wu2018mixed} & \textgreater{}2/4 & 68.65 \\
 & \text{AutoQNN} & 2.13/2.49 & \underline{68.84} \\
{ResNet18} & AutoQB-QBN~\cite{AutoQB} & 3.12/3.29 & 67.63 \\
 & AutoQB-BBN~\cite{AutoQB} & 3.06/3.27 & 63.41 \\
 & \text{AutoQNN} & 2.94/3.07 & \underline{69.88} \\
 
{MobileNets} & HAQ~\cite{wang2019haq} & 2.16/32 & 57.14 \\
 & \text{AutoQNN} & 2.28/32 & \underline{69.88} \\
 
{MobileNetV2} & HAQ & 2.27/32 & 66.75 \\
 & \text{AutoQNN} & 2.30/32 & \underline{69.77} \\

{ResNet50} & HAQ & 2.06/32 & 70.63 \\
 & \gcc{\text{BSQ}~\cite{yang2021bsq}} & \gcc{2.30/4} & \gcc{\underline{75.16}} \\
 & AutoQNN & 2.27/32 & 74.76 \\

{InceptionV3} & HAWQ~\cite{HAWQ} & 2.60/4 & 75.52 \\
 & HAWQ-V2~\cite{HAWQV2} & 2.61/4 & 75.98 \\
 & \gcc{BSQ} & \gcc{2.48/6} & \gcc{75.90} \\
 & \gcc{\text{BSQ}} & \gcc{2.81/6} & \gcc{\underline{76.60}} \\
 & AutoQNN & 2.69/4.00 & 76.57\\
\Xhline{1.5pt}
\end{tabular}\flushleft\vspace{-8pt}
\gccpf{Note: }{W/A indicates the average bitwidth for weight and activation, respectively.
}
\gccpf{The best results are \underline{underlined}.}
\end{table}

\subsubsection{Comparison With Mixed-Precision Schemes} 
In this experiment, we verify that the proposed AutoQNN can outperform \sArt~mixed-precision schemes and gain higher accuracy in DNN quantization.
According to the results in~\subsecref{sec:comparing_with_fixed_prcision}, we employ PotQ for weight quantization and ClipQ for activation quantization, respectively.
We seek reasonable mixed-precision policies by QPL and compare them with the mixed-precision schemes under the same average bitwidth condition.

\cumparagraph{Comparison results}

The comparison results are shown in \tabref{tab:comparing_with_mixedp}.       
Compared with MixedP~\cite{wu2018mixed} on ResNet18, AutoQNN achieves a higher accuracy of \text{68.84\%} with the lower average bitwidths of \text{2.16/2.49} for weight and activation quantization. 
AutoQNN outperforms AutoQB-QBN~\cite{AutoQB} and AutoQB-BBN~\cite{AutoQB} by \text{2.25\%} and \text{6.47\%} improvements in model accuracy, respectively, under even lower average bitwidths of \text{2.94/3.07}.
When we employ AutoQNN to quantize models, including MobileNets, MobileNetV2, and ResNet50, into the same size as that in HAQ~\cite{wang2019haq}, we can gain accuracy improvements by \text{12.74\%}, \text{3.02\%}, and \text{4.13\%}, respectively.
Besides, there are \text{1.05\%} and \text{0.59\%} accuracy improvements on InceptionV3 by AutoQNN, compared with HAWQ~\cite{HAWQ} and HAWQ-V2~\cite{HAWQV2}, respectively.
\gcc{On InceptionV3, AutoQNN with \text{2.69/4} achieves higher accuracy (76.57\% vs. 75.90\%) using even lower activation bitwidth compared with BSQ with \text{2.48/6}, and also uses lower weight and activation bitwidth to achieve similar accuracy (76.57\% vs. 76.60\%) to BSQ with \text{2.81/6}.}
\gcc{The top1 accuracy of BSQ~\cite{yang2021bsq} with \text{2.3/4} on ResNet50 over ImageNet achieves 75.16\%, which slightly exceeds the accuracy result $74.76\%$ of AutoQNN with \text{2.27/32}.
The reason is that the referenced pre-trained ResNet50 in PyTorch has relatively higher accuracy than that in Keras.
The ResNet50 from the Keras community has an accuracy of $74.90\%$, but that in the PyTorch community is up to $76.15\%$.
According to the referenced accuracy, the accuracy drop of AutoQNN is only 0.14\%, while that of BSQ is up to $0.99\%$.
}
These comprehensive comparison results highlight that AutoQNN can obtain better mixed-precision policies for various mainstream architectures and surpass the \sArt~methods. 

\begin{table*}[!ht]
    \caption{\gccpf{Accuracy Results of LSTM Model on THCUNews} \label{tab:cnews_classification}}
    \centering
    \setlength{\tabcolsep}{13.8pt}
    \begin{tabular}{lcccccc}
    \Xhline{1.5pt}
    \gcc{Method} &\gccpf{ Weight }& \gccpf{Activation} & \gccpf{Average Bits} & \gccpf{Average Bits} & \gcc{Accuracy} \\ 
     &\gccpf{Quantization }& \gccpf{Quantization} & \gccpf{of Weights} & \gccpf{of Activation} & \\ \Xhline{1.5pt}
    \gccpf{Full-Precision} & - & - & \gcc{32} & \gcc{32} & \gcc{95.52} \\ 
    {\gcc{\text{AutoQNN}}} & {\gcc{ZoomQ}} & {\gcc{PotQ}} & \gcc{2.00} & \gcc{2.01} & \gcc{94.53 }\\ 
     &  &  & \gcc{2.00} & \gcc{3.01} & \gcc{95.03} \\ 
     &  &  & \gcc{3.01} & \gcc{3.02} & \gcc{95.46} \\ \Xhline{1.5pt}
    \end{tabular}
\end{table*}
\begin{table*}[!ht]
    \centering
    \caption{\gccpf{PPW Results of Different Quantization Methods on PTB}}
    \label{tab:ptb_prediction}
    \setlength{\tabcolsep}{15pt}
    \begin{tabular}{lccccc}
    \Xhline{1.5pt}
    \gcc{Method} & \gcc{Weight} & \gccpf{Activation} & \gccpf{Average Bits} & \gccpf{Average Bits} & \gcc{PPW} \\
      & \gccpf{Quantization} & \gccpf{Quantization} & \gccpf{of Weights} & \gccpf{of Activation} &  \\ \Xhline{1.5pt}
    \gcc{EffectiveQ~\cite{he2016effectivequantization}} & - & - & \gcc{2} & \gcc{2} & \gcc{152} \\
    \gcc{EffectiveQ} & - & - & \gcc{2} & \gcc{3} & \gcc{142} \\ 
    \gcc{EffectiveQ} & - & - & \gcc{3} & \gcc{3} & \gcc{120} \\
    \gcc{QNNs~\cite{hubara2017quantizednn}} & - & - & \gcc{2} & \gcc{3} & \gcc{220} \\ 
    \gcc{LP-RNNs~\cite{kapur2017low}} & - & - & \gcc{2} & \gcc{2} & \gcc{152.2} \\
    \gcc{BalancedQ~\cite{zhou2017balanced}} & - & - & \gcc{2} & \gcc{2} & \gcc{126} \\ 
    \gcc{BalancedQ} & - & - & \gcc{2} & \gcc{3} & \gcc{123} \\ 
    \gcc{HitNet~\cite{peiqi2018hitnet}} & \gcc{TTQ} & \gcc{BTQ} & \gcc{2} & \gcc{2} & \gcc{\text{110.3}} \\ 
    \gcc{\text{AutoQNN}} & \gcc{ResQ} & \gcc{ClipQ} & \gcc{2.48} & \gcc{2.28} & \gcc{116.7} \\ \Xhline{1.5pt}
    \end{tabular}
\end{table*}
\subsection{\gcc{Evaluation on LSTM}}
\gcc{We conduct two long-short-term-memory (LSTM) \cite{hochreiter1997lstm} experiments in this \gccpf{subsection} to verify the effectiveness of AutoQNN on natural language processing (NLP) applications. 
We employ AutoQNN to search different quantization policies for four weight matrices and one output state in LSTMs, as did in HitNet~\cite{peiqi2018hitnet}.
The experimental results show that AutoQNN can find the appropriate quantizing strategies for the different weights and activation in LSTMs.}

\subsubsection{\gcc{Experiments on Text Classification}}
\gcc{We first evaluate AutoQNN on the text classification task over the subset of THCUNews dataset\footnote{https://github.com/thunlp/THUCTC. Oct. 2022.},
which contains \gccpf{50k news} for 10 categories. 
We use the model with one word embedding layer, one LSTM layer (with 512 hidden units), and two fully-connected layers (with 256 and 128 hidden units, respectively) as \gccpf{an evaluated model}, and employ accuracy to measure model performance.
We quantize the weights and activation of the embedding, LSTM, and fully-connected layers in \gccpf{the evaluated model}.}

\gcc{The experimental results are shown in \tabref{tab:cnews_classification}.
AutoQNN finds that ZoomQ and PotQ are the best quantizing schemes for weight and activation quantization, respectively.
The accuracy of the quantized model using 2-bit weights and 2-bit activation achieves $94.53\%$, slightly lower than the full-precision result by $0.99\%$.
When increasing the quantizing bitwidth to 3, the model accuracy achieves $95.46\%$, which is very close to the accuracy of the full-precision model.
This experiment shows that AutoQNN can be applied to text classification tasks to preserve the accuracy of quantized recurrent neural networks (RNNs).}

\begin{figure*}[!t]
    \centering
    \includegraphics[width=1\textwidth]{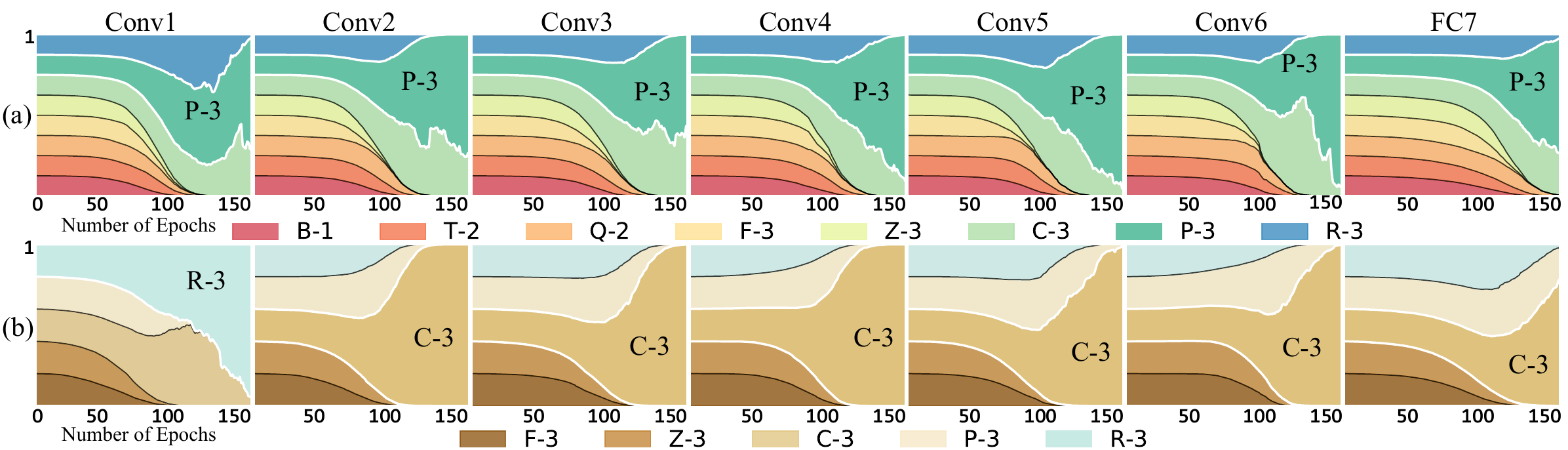}
    \caption{\label{fig:qssf_scheme_details_on_cifar10}The changes of sampling probabilities of candidate schemes in the training phase of VGG-like. The \emph{x}-axis is the number of training epochs(total number of 150 epochs) and the \emph{y}-axis is the sampling probabilities of schemes (sum of the probabilities is 1). \gccpf{(a) Quantizing scheme search for the weights of different layers. (b) Quantizing scheme search for the activation of different layers.}}
\end{figure*}
\subsubsection{\gcc{Experiments on Penn TreeBank}}
\gcc{We further evaluate AutoQNN on sequence prediction task over the penn treebank (PTB) dataset~\cite{taylor2003ptb}, which contains \gccpf{10k} unique words. 
We use an open model implementation for evaluation, and its source codes can be found here\footnote{https://github.com/adventuresinML/adventures-in-ml-code. Oct. 2022.}.
For a fair comparison, we modify the open model and use one embedding layer with \gccpf{300 outputs} and one LSTM layer with 300 hidden units, as did in \cite{he2016effectivequantization}.
We train the full-precision model for 100 epochs with \gccpf{adam optimizer~\cite{kingma2015adam}} and a learning rate of 1e-3.
Then, we adopt AutoQNN to search for the best quantizing strategy for the full-precision model and quantize the weights and activation of its embedding and LSTM layers.
Specifically, we employ the coarse-grained QSS to seek a shared quantizing scheme and use QPL to learn a mixed-precision policy.
Model performance is measured in perplexity per word (PPW) metric, as used in \cite{peiqi2018hitnet,he2016effectivequantization,hubara2017quantizednn,zhou2017balanced}.}

\gcc{The comparison results are shown in \tabref{tab:ptb_prediction}. 
AutoQNN employs the 2.48-bit ResQ for weight quantization and the 2.28-bit ClipQ for activation quantization, and achieves a competitive PPW result, outperforming previous \gccpf{methods}, including EffectiveQ~\cite{he2016effectivequantization}, QNNs~\cite{hubara2017quantizednn}, LP-RNNs~\cite{kapur2017low}, and BalancedQ~\cite{zhou2017balanced}.
The result of AutoQNN is slightly higher than that of HitNet~\cite{peiqi2018hitnet}, because we use an inaccurate full-precision model with a compressed embedding layer.
Another reason is that the candidate schemes in AutoQNN are designed for CNNs and are not adapted to RNNs. 
For example, the activation function applied in RNNs is $tanh$, and its output range is $(-1,$ $1)$. 
In contrast, the activation range of CNN is usually $[0,$ $+\infty)$. 
Therefore, applying truncation for RNN's outputs in quantization is harmful to model performance. 
Nevertheless, AutoQNN can still find the best quantizing strategies for RNNs to preserve PPW.}

\begin{figure}[H]
    \centering
    \includegraphics[width=1\columnwidth]{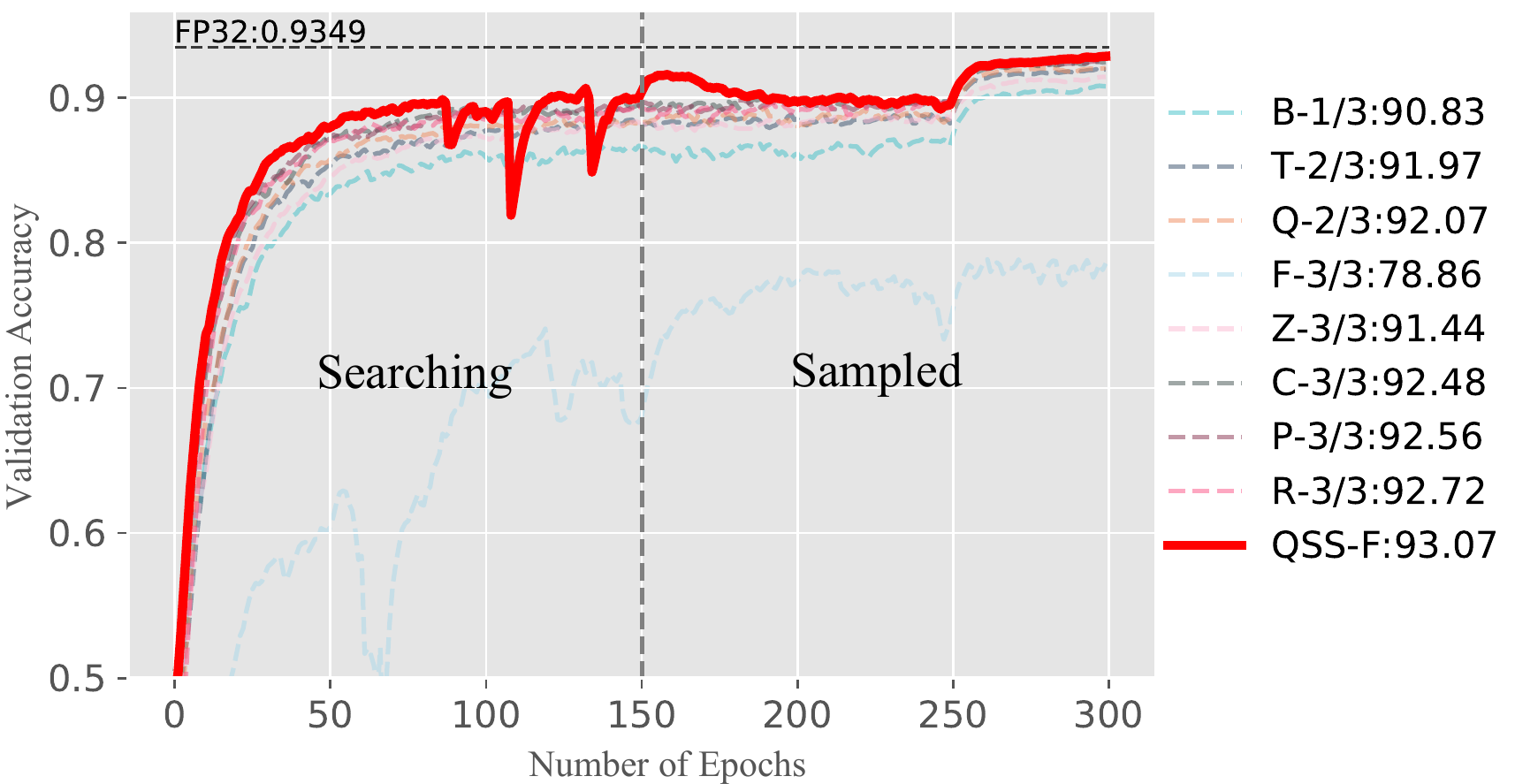}
    \caption{\label{fig:qssf_accuracy_on_cifar10}The validation accuracy curves in model training. QSS-F denotes the result of the model quantized with the layer-wised search scheme.}
\end{figure}
\subsection{Ablation Study}
In this subsection, we perform two ablation studies including QSS validation and QPL validation. 

\subsubsection{QSS Validation}
We evaluate the attainable accuracy of quantized models to verify that QSS can find desirable quantizing schemes for DNNs\footnote{More experimental results can be found in our supplementary materials.}.
To do this, we take all candidate schemes as baselines and conduct two experiments: layer-wised search and model-wised search.

\cumparagraph{Settings}

The widely verified Cifar10~\cite{krizhevsky2009learning} and VGG-like~\cite{simonyan2014vgg} are used in this experiment. The data augmentation of Cifar10 in~\cite{lee2015deeply} is employed.
To highlight the discrimination of results, we fix the quantizing bitwidth as 3 for the multi-bit schemes such as ClipQ and PotQ. 
We train VGG-like on Cifar10 for 300 epochs with an initial learning rate of 0.1, and decay the learning rate to 0.01 and 0.001 at 250 and 290 epochs, respectively.

\cumparagraph{Layer-Wised Search}

We present layer-wised search processes in \figref{fig:qssf_scheme_details_on_cifar10}, which draws the changes of sampling probabilities of different candidate schemes during 150 training epochs.
All the schemes have the same sampling probabilities at the beginning of a training phase.
The sum of the probabilities at each epoch equals 1.
For the weight quantization of the 7 layers of VGG-like, the search processes of different layers tend to be similar, i.e., the probability of P-3 gradually grows with the training epoch increasing, while that of other schemes decreases.
The results imply that P-3 contributes less training loss during model training than other schemes.
Therefore, QSS finds that P-3 is the desirable quantizing scheme among all candidate schemes.
Similarly, for the activation quantization of VGG-like, QSS has found that R-3 owns the highest sampling probability at the first layer because R-3 can reserve the features in the first layer well. 
Besides, C-3 achieves the highest sampling probabilities at the rest 6 layers since C-3 can eliminate outliers and contribute robust training results.
Finally, QSS finds the desirable quantizing schemes of \gccpf{\{P-3, P-3, P-3, P-3, P-3, P-3, P-3\}} and \gccpf{\{R-3, C-3, C-3, C-3, C-3, C-3, C-3\}} for the weight and activation quantization of VGG-like, respectively.

Next, we generate a quantized architecture with the found schemes above and fine tune it for another 150 epochs to converge, as described in~\algref{alg:model_training}. 
The accuracy comparison is shown in \figref{fig:qssf_accuracy_on_cifar10}, QSS is denoted as QSS-F.
B-1/3 denotes employing 1-bit Binary for weight quantization and 3-bit ClipQ for activation quantization, respectively. 
The accuracy of QSS-F achieves \text{93.07\%}, which constantly outperforms that of candidate schemes by \text{2.70\%} on average, and only incurs \text{0.42\%} accuracy degradation compared with the full-precision result (denoted as FP32).
The result verifies that QSS is able to seek desirable quantizing schemes for different layers to gain high accuracy.

\begin{figure}[H]
    \centering
    \subfloat[]
    {\label{fig:QSS_C_W_on_cifar10}
    \includegraphics[width=0.522\columnwidth]{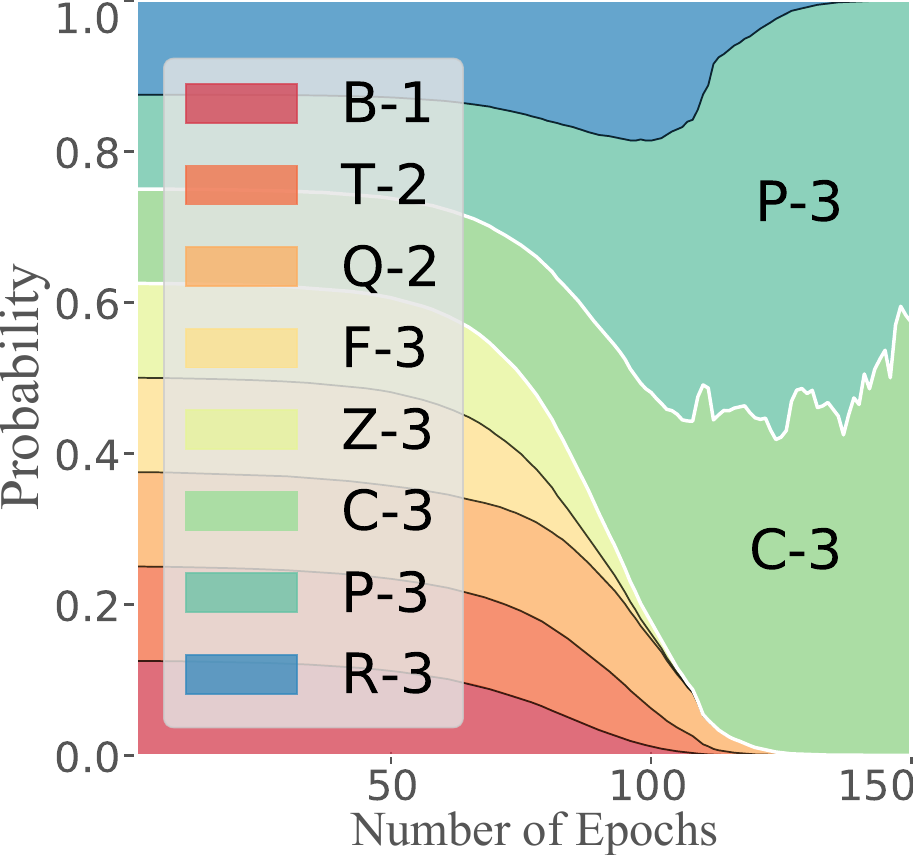}}
    \subfloat[]
    {
    \label{fig:QSS_C_A_on_cifar10}
    \includegraphics[width=0.465\columnwidth]{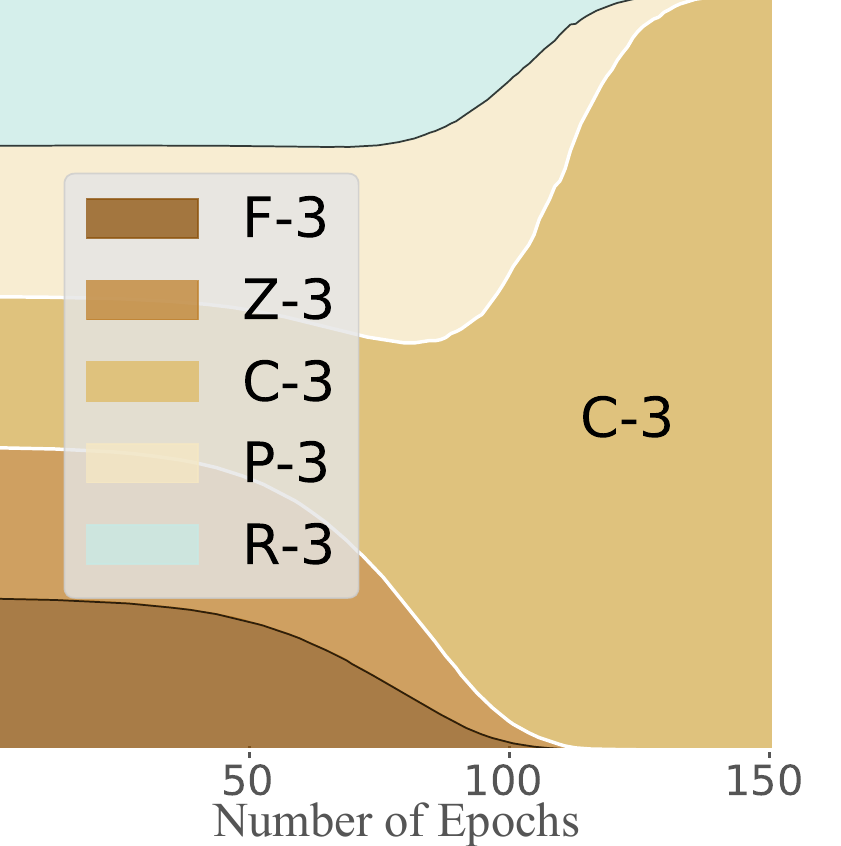}}
    \caption{The changes of shared sampling probabilities in model-wised search.
    \gccpf{(a) Quantizing scheme search for the weights in VGG-like. (b) Quantizing scheme search for the activation in VGG-like.}}
    \label{fig:QSS_C_on_cifar10}
\end{figure}
\renewcommand{\arraystretch}{1.1}
\begin{table}[H]
\centering
\caption{\label{tab:qss_c_accuracy}\gccpf{Accuracy Comparison in Model-Wised Search (C-3/C-3 and P-3/C-3)}}
\setlength{\tabcolsep}{11pt}
\begin{tabular}{lcc}\Xhline{1.5pt}
\gccpf{Name} & \gccpf{Method} & \gccpf{Accuracy} \\ \Xhline{1.5pt}
\gccpf{Normal Quantization} & B-1/3 & 90.83 \\
& T-2/3 & 91.97 \\
& Q-2/3 & 92.07 \\
& F-3/3 & 78.86 \\
& Z-3/3 & 91.44 \\
& P-3/3 & 92.56 \\
& R-3/3 & 92.72 \\
\gccpf{Model-Wised Search}& C-3/C-3 & 92.48 \\
& P-3/C-3 & \underline{92.90} 
\\\Xhline{1.5pt}
\end{tabular}\flushleft\vspace{-8pt}
\gccpf{Note: The best results are underlined.}
\end{table}
\cumparagraph{Model-Wised Search}

We employ the coarse-grained QSS to find shared quantizing schemes for all layers of \gccpf{VGG-like}.
Specifically, respecting that the distributions of weights and activation are usually diverse, we seek two shared schemes for the weight and activation quantization in this experiment, respectively.
\begin{figure*}[!ht]
    \flushleft
    \subfloat[]{
    \label{fig:QPL_on_cifar10_weight_bits}
    \includegraphics[width=0.31\textwidth]{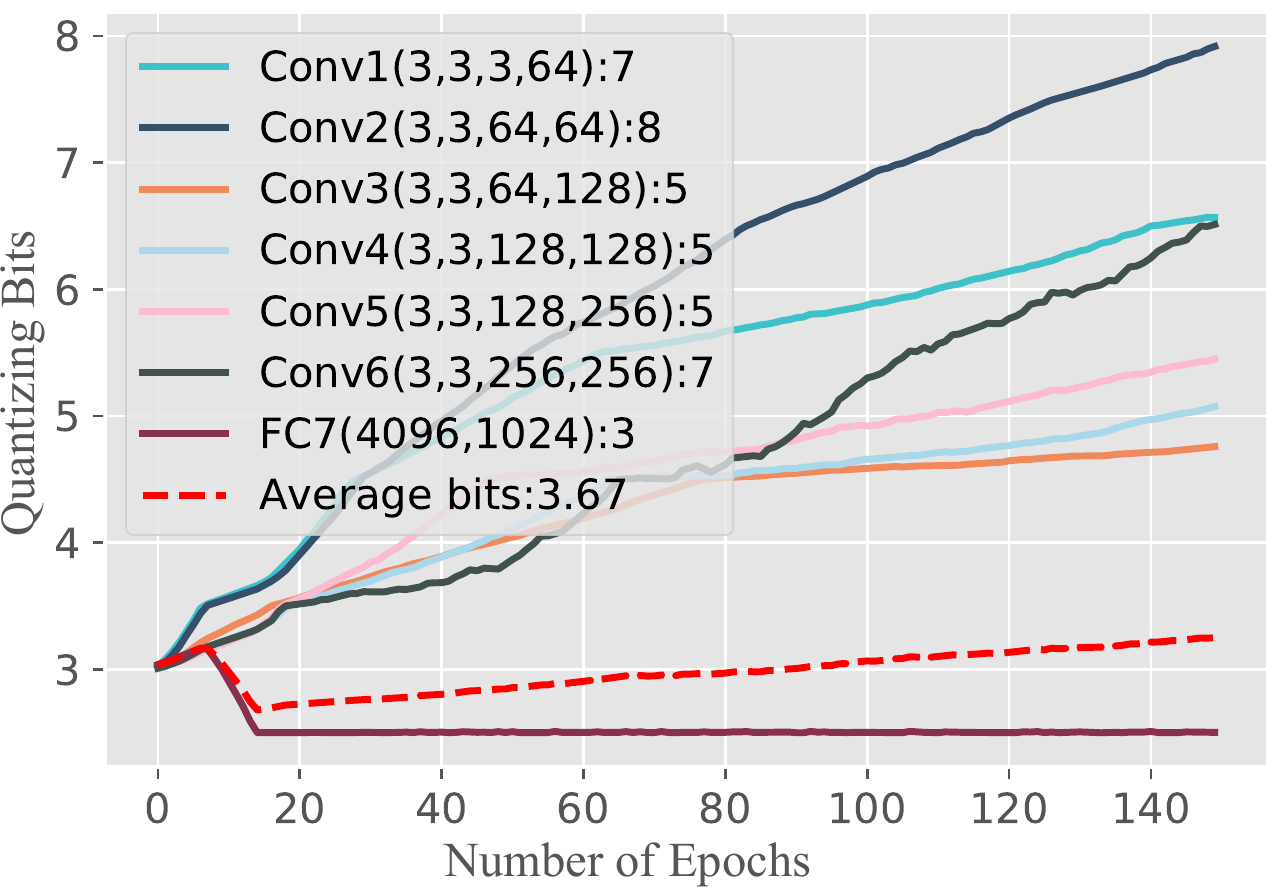}}
    \subfloat[]{
    \label{fig:QPL_on_cifar10_activation_bits}
    \includegraphics[width=0.31\textwidth]{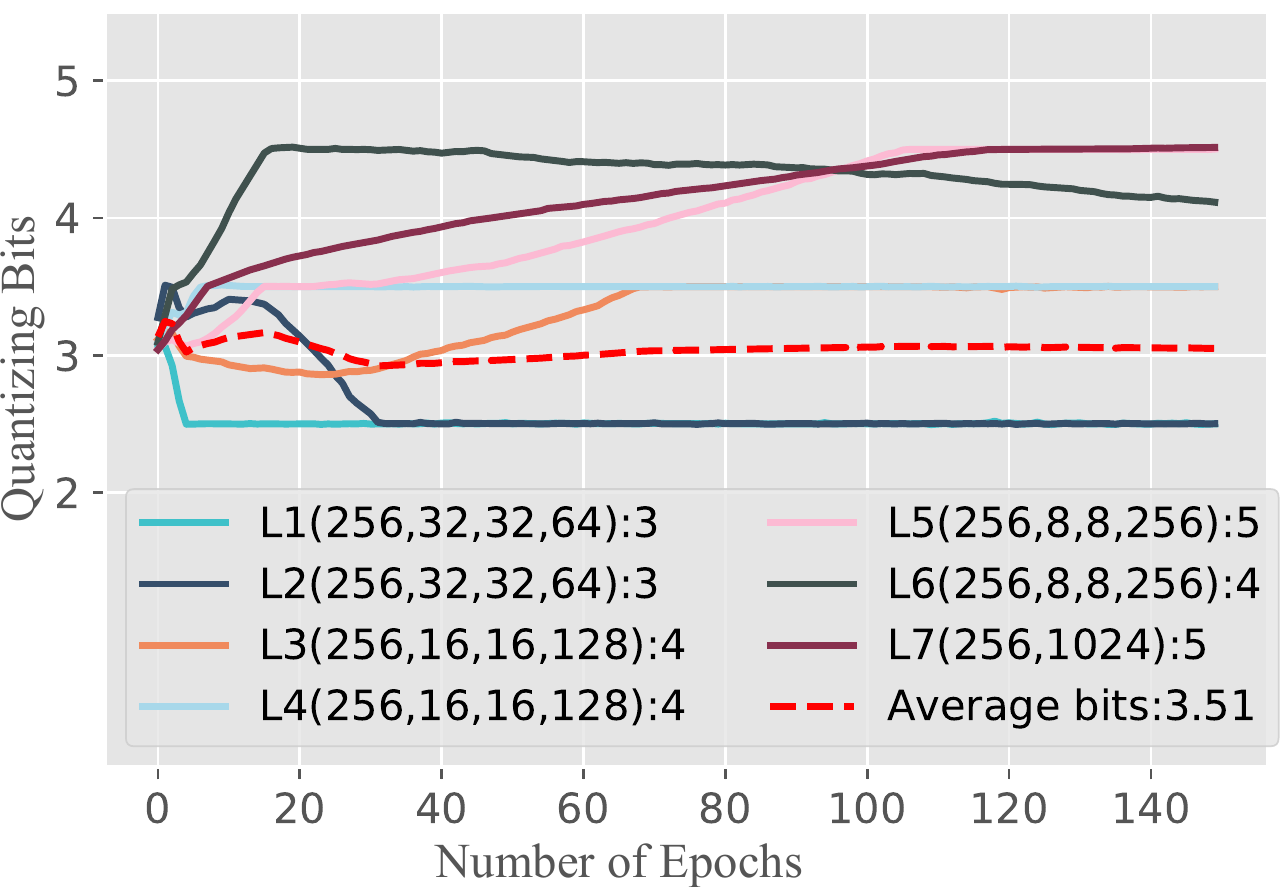}}
    \subfloat[]{
    \label{fig:QPL_on_cifar10_accuracy}
    \includegraphics[width=0.33\textwidth]{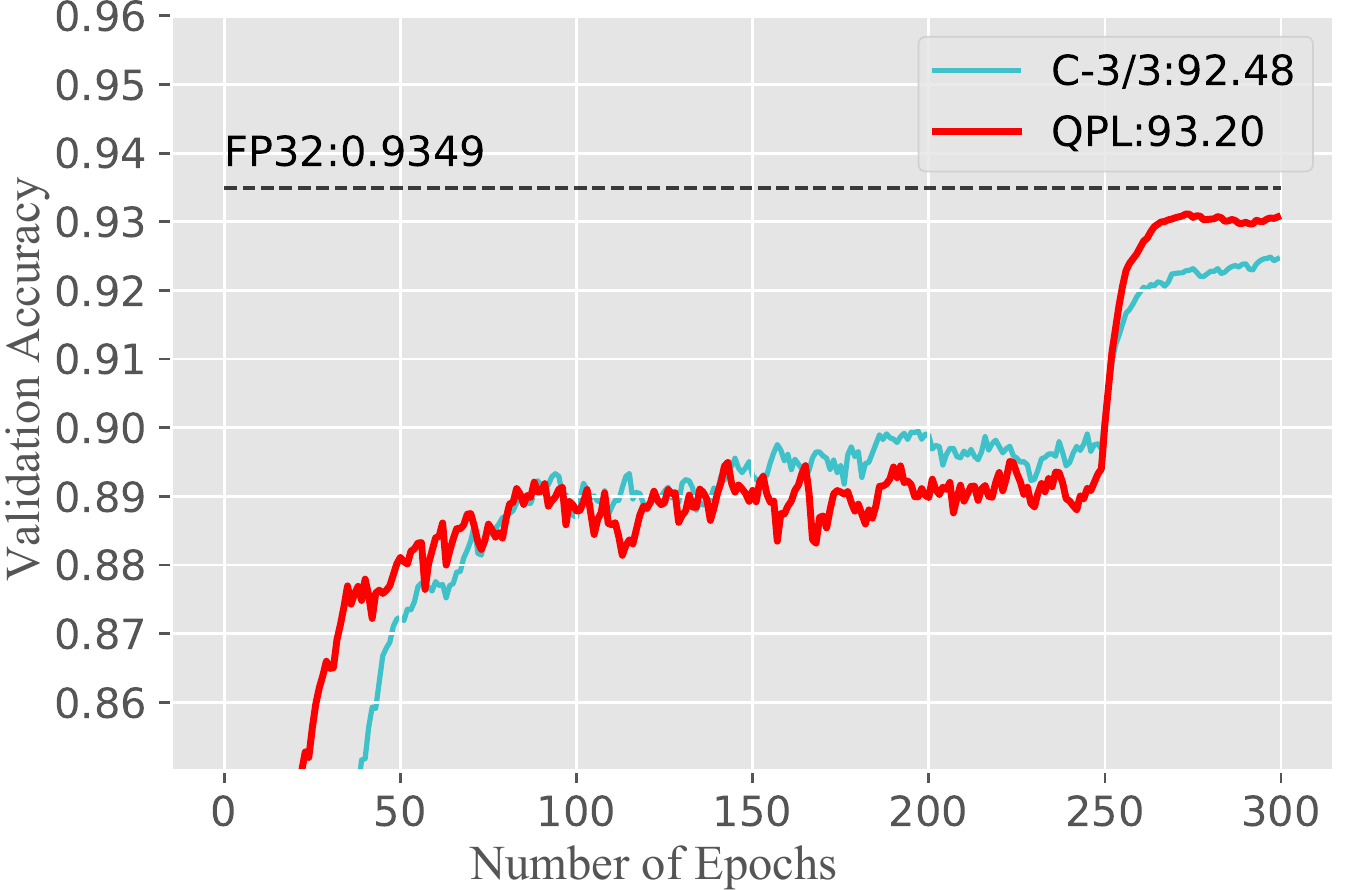}}
    \caption{\label{fig:qpl_learning_results}The learning process of QPL and the final validation accuracy comparison of quantized VGG-like.
    \gccpf{(a) The bit changes of weight quantization. (b) The bit changes of activation quantization. (c) Accuracy results.}}
    \label{fig:QPL_on_cifar10_results}
\end{figure*}
The search process for weight quantization is shown in \gccpf{\figref{fig:QSS_C_on_cifar10}(a)}, and that for activation quantization is shown in \gccpf{\figref{fig:QSS_C_on_cifar10}(b)}.
\gccpf{Similar to} the layer-wised results above, QSS finds that C-3 and P-3 are the desirable quantizing schemes for weight quantization, and C-3 is the desirable one for activation quantization. 
Therefore, we take C-3/C-3 and P-3/C-3 as two desirable quantizing strategies. 
C-3/C-3 employs C-3 for both weight and activation quantization.
P-3/C-3 adopts P-3 for weight quantization and uses C-3 for activation quantization. 

Since the shared quantizing schemes may not meet the requirements of some layers inevitably, model-wised search sacrifices some accuracy to obtain the unified quantizing schemes.
However, QSS still presents competitive results compared with baselines.
As shown in \tabref{tab:qss_c_accuracy}, 
the accuracy of C-3/C-3 achieves \text{92.48\%}, which is only slightly lower than the results of P-3/3 and R-3/3. 
The accuracy of P-3/C-3 achieves \text{92.90\%}, which outperforms that of P-3/3 and R-3/3 by \text{0.34\%} and \text{0.18\%}, respectively. 
These results demonstrate that QSS is able to find expected quantizing schemes from a candidate set efficiently, thus improving the accuracy of quantized models without requiring more bits and avoiding heavy manual workloads.

\subsubsection{QPL Validation}
In this subsection, we verify that QPL can find reasonable mixed-precision policies for VGG-like and reduce the performance degradation of the quantized model within limited model size and memory footprint.


We take ClipQ as the quantizing scheme for verifying QPL, since clip quantization is the most widely used scheme.
We still utilize Cifar10 and VGG-like for evaluations.
\gcc{In addition, we set a target expected bit of 3 for precision loss in this experiment to highlight comparison results.}

We train 150 epochs to learn optimal mixed-precision policies for VGG-like.
\gccpf{\figref{fig:QPL_on_cifar10_results}(a)} presents the quantizing bit changes of weight quantization with the number of epochs.
We finally obtain the mixed-precision policy of \gccpf{\{7, 8, 6, 6, 7, 5, 2\}} for the weight quantization of different layers, and the average bitwidth of this policy is \text{2.82}.
There are $4096\times1024$ parameters in the full-connected layer \text{FC7}, which occupy over 90\% of the VGG-like parameters and have a lot of redundancy.
Therefore, QPL obtains the bitwidth of 2 for \text{FC7} to balance the classification and precision losses. 
The convolutional layers use a small number of parameters to extract features and have less redundancy. 
Consequently, QPL learns high bitwidths to reduce the classification loss, such as the 7 and 8 bits for Conv1 and Conv2, respectively, as shown in \gccpf{\figref{fig:QPL_on_cifar10_results}(a)}.
The mixed-precision learning process of activation quantization is shown in \gccpf{\figref{fig:QPL_on_cifar10_results}(b)}.
QPL learns the mixed-precision policy of \gccpf{\{2, 3, 3, 4, 5, 4, 6\}} and the average bitwidth of this policy is \text{3.08}.
The accuracy comparison is shown in \gccpf{\figref{fig:QPL_on_cifar10_results}(c)}. 
Compared with ClipQ (denoted as C-3/3), the accuracy of the learned mixed-precision policy (denoted as C-2.82/3.08) achieves~\text{93.03\%}, which outperforms C-3/3 by \text{0.55\%}.

The results above demonstrate that QPL is able to learn relatively optimal mixed-precision policies to balance the classification and precision losses of DNNs. 
QPL employs the classification loss to reduce the model redundancy and proposes the precision loss $\overline{L}$ in \eqref{eq:average_bits} to constrain the model size and memory footprint, thus improving the accuracy and efficiency of quantized DNNs.

\section{Conclusions}\label{sec:con}
In this paper,
we proposed AutoQNN, an end-to-end framework aiming at automatically quantizing neural networks. 
Differing from manual-participated heuristic explorations with heavy workloads of domain experts, AutoQNN could efficiently explore the search space of automatic quantization and provide appropriate quantizing strategies for arbitrary DNN architectures. 
It automatically sought desirable quantizing schemes and learned relatively optimal mixed-precision policies for efficiently compressing DNNs.
Compared with full-precision models, the quantized models using AutoQNN achieved competitive classification accuracy with much smaller model size and memory footprint.
Compared with \sArt~competitors, the comprehensive evaluations on AlexNet and ResNet18 demonstrated that AutoQNN obtained accuracy improvements by up to 1.65\% and 1.74\%, respectively.
\vspace{2mm}
\bibliographystyle{JCST}
\bibliography{references}

\begin{thebibliography}{10}

\bibitem{williams2009roofline}
Williams S, Waterman A, Patterson D.
\newblock Roofline: an insightful visual performance model for multicore
  architectures.
\newblock {\em Communications of the ACM}, 2009, 52(4):65--76.

\bibitem{hubara2016binarized}
Hubara I, Courbariaux M, Soudry D, El{-}Yaniv R, Bengio Y.
\newblock Binarized neural networks.
\newblock In {\em {NIPS}}, 2016, pp. 4107--4115.

\bibitem{TWNs}
Li F, Liu B.
\newblock Ternary weight networks.
\newblock {\em CoRR}, 2016, abs/1605.04711.

\bibitem{zhou2016dorefa}
Zhou S, Ni Z, Zhou X, Wen H, Wu Y, Zou Y.
\newblock Dorefa-net: Training low bitwidth convolutional neural networks with
  low bitwidth gradients.
\newblock {\em CoRR}, 2016, abs/1606.06160.

\bibitem{Ternaryconnect}
Lin Z, Courbariaux M, Memisevic R, Bengio Y.
\newblock Neural networks with few multiplications.
\newblock In {\em {ICLR}}, 2016.

\bibitem{VecQ}
Gong C, Chen Y, Lu Y, Li T, Hao C, Chen D.
\newblock Vecq: Minimal loss dnn model compression with vectorized weight
  quantization.
\newblock {\em IEEE Transactions on Computers}, 2020, 70(5):696--710.
\newblock DOI: 10.1109/TC.2020.2995593.

\bibitem{cheng2019uL2Q}
Gong C, Li T, Lu Y, Hao C, Zhang X, Chen D, Chen Y.
\newblock {\(\mathrm{\mu}\)}l2q: An ultra-low loss quantization method for
  {DNN} compression.
\newblock In {\em IJCNN}, 2019, pp. 1--8.
\newblock DOI: 10.1109/ijcnn.2019.8851699.

\bibitem{HAWQ}
Dong Z, Yao Z, Gholami A, Mahoney M~W, Keutzer K.
\newblock {HAWQ:} hessian aware quantization of neural networks with
  mixed-precision.
\newblock In {\em {IEEE ICCV}}, 2019, pp. 293--302.
\newblock DOI: 10.1109/iccv.2019.00038.

\bibitem{HAWQV2}
Dong Z, Yao Z, Arfeen D, Gholami A, Mahoney M~W, Keutzer K.
\newblock {HAWQ-V2:} hessian aware trace-weighted quantization of neural
  networks.
\newblock In {\em {NIPS}}, 2020.

\bibitem{wang2019haq}
Wang K, Liu Z, Lin Y, Lin J, Han S.
\newblock Haq: Hardware-aware automated quantization with mixed precision.
\newblock In {\em {IEEE CVPR}}, 2019, pp. 8612--8620.
\newblock DOI: 10.1109/cvpr.2019.00881.

\bibitem{lin2016fixed}
Lin D~D, Talathi S~S, Annapureddy V~S.
\newblock Fixed point quantization of deep convolutional networks.
\newblock In {\em {ICML}}, volume~48 of {\em {JMLR} Workshop and Conference
  Proceedings}, 2016, pp. 2849--2858.

\bibitem{wu2018mixed}
Wu B, Wang Y, Zhang P, Tian Y, Vajda P, Keutzer K.
\newblock Mixed precision quantization of convnets via differentiable neural
  architecture search.
\newblock {\em CoRR}, 2018, abs/1812.00090.

\bibitem{AutoQB}
Lou Q, Liu L, Kim M, Jiang L.
\newblock Autoqb: Automl for network quantization and binarization on mobile
  devices.
\newblock {\em CoRR}, 2019, abs/1902.05690.

\bibitem{zhu2016TTQ}
Zhu C, Han S, Mao H, Dally W~J.
\newblock Trained ternary quantization.
\newblock {\em CoRR}, 2016, abs/1612.01064.

\bibitem{ENN2017}
Leng C, Dou Z, Li H, Zhu S, Jin R.
\newblock Extremely low bit neural network: Squeeze the last bit out with
  {ADMM}.
\newblock In {\em {AAAI}}, 2018, pp. 3466--3473.
\newblock DOI: 10.1609/aaai.v32i1.11713.

\bibitem{BMobileES}
Phan H, Liu Z, Huynh D, Savvides M, Cheng K, Shen Z.
\newblock Binarizing mobilenet via evolution-based searching.
\newblock In {\em {IEEE CVPR}}, 2020, pp. 13417--13426.
\newblock DOI: 10.1109/CVPR42600.2020.01343.

\bibitem{Binaryconnect}
Courbariaux M, Bengio Y, David J.
\newblock Binaryconnect: Training deep neural networks with binary weights
  during propagations.
\newblock In {\em {NIPS}}, 2015, pp. 3123--3131.

\bibitem{rastegari2016xnor}
Rastegari M, Ordonez V, Redmon J, Farhadi A.
\newblock Xnor-net: Imagenet classification using binary convolutional neural
  networks.
\newblock In {\em {ECCV}}, volume 9908, 2016, pp. 525--542.
\newblock DOI: 10.1007/978-3-319-46493-0\_32.

\bibitem{alemdar2017ternary}
Alemdar H, Leroy V, Prost{-}Boucle A, P{\'{e}}trot F.
\newblock Ternary neural networks for resource-efficient {AI} applications.
\newblock In {\em IJCNN}, 2017, pp. 2547--2554.
\newblock DOI: 10.1109/ijcnn.2017.7966166.

\bibitem{jin2018sparse}
Jin C, Sun H, Kimura S.
\newblock Sparse ternary connect: Convolutional neural networks using
  ternarized weights with enhanced sparsity.
\newblock In {\em ASP-DAC}, 2018, pp. 190--195.
\newblock DOI: 10.1109/aspdac.2018.8297304.

\bibitem{gysel2016hardware}
Gysel P.
\newblock Ristretto: Hardware-oriented approximation of convolutional neural
  networks.
\newblock {\em CoRR}, 2016, abs/1605.06402.

\bibitem{chen2019tdla}
{Chen} Y, {Zhang} K, {Gong} C, {Hao} C, {Zhang} X, {Li} T, {Chen} D.
\newblock {T-DLA}: An open-source deep learning accelerator for ternarized dnn
  models on embedded fpga.
\newblock In {\em ISVLSI}, 2019, pp. 13--18.
\newblock DOI: 10.1109/isvlsi.2019.00012.

\bibitem{APoT}
Li Y, Dong X, Wang W.
\newblock Additive powers-of-two quantization: {A} non-uniform discretization
  for neural networks.
\newblock {\em CoRR}, 2019, abs/1909.13144.

\bibitem{TSQ2018}
Wang P, Hu Q, Zhang Y, Zhang C, Liu Y, Cheng J.
\newblock Two-step quantization for low-bit neural networks.
\newblock In {\em {IEEE CVPR}}, 2018, pp. 4376--4384.
\newblock DOI: 10.1109/cvpr.2018.00460.

\bibitem{QIL}
Jung S, Son C, Lee S, Son J, Han J, Kwak Y, Hwang S~J, Choi C.
\newblock Learning to quantize deep networks by optimizing quantization
  intervals with task loss.
\newblock In {\em {IEEE CVPR}}, 2019, pp. 4350--4359.
\newblock DOI: 10.1109/CVPR.2019.00448.

\bibitem{PACT}
Choi J, Wang Z, Venkataramani S, Chuang P~I, Srinivasan V, Gopalakrishnan K.
\newblock {PACT:} parameterized clipping activation for quantized neural
  networks.
\newblock {\em CoRR}, 2018, abs/1805.06085.

\bibitem{LogQ}
Miyashita D, Lee E~H, Murmann B.
\newblock Convolutional neural networks using logarithmic data representation.
\newblock {\em CoRR}, 2016, abs/1603.01025.

\bibitem{INQ2017}
Zhou A, Yao A, Guo Y, Xu L, Chen Y.
\newblock Incremental network quantization: Towards lossless cnns with
  low-precision weights.
\newblock {\em CoRR}, 2017, abs/1702.03044.

\bibitem{ghasemzadeh2018rebnet}
Ghasemzadeh M, Samragh M, Koushanfar F.
\newblock Rebnet: Residual binarized neural network.
\newblock In {\em FCCM}, 2018, pp. 57--64.
\newblock DOI: 10.1109/fccm.2018.00018.

\bibitem{Rebinary}
Li Z, Ni B, Zhang W, Yang X, Gao W.
\newblock Performance guaranteed network acceleration via high-order residual
  quantization.
\newblock In {\em {IEEE ICCV}}, 2017, pp. 2603--2611.
\newblock DOI: 10.1109/iccv.2017.282.

\bibitem{he2016resnet}
He K, Zhang X, Ren S, Sun J.
\newblock Deep residual learning for image recognition.
\newblock In {\em {IEEE CVPR}}, 2016, pp. 770--778.
\newblock DOI: 10.1109/cvpr.2016.90.

\bibitem{residualquantization2021tom}
Li Z, Ni B, Li T, Yang X, Zhang W, Gao W.
\newblock Residual quantization for low bit-width neural networks.
\newblock {\em IEEE Transactions on Multimedia}, 2021, pp. 1--1.
\newblock DOI: 10.1109/TMM.2021.3124095.

\bibitem{LQ-Nets}
Zhang D, Yang J, Ye D, Hua G.
\newblock Lq-nets: Learned quantization for highly accurate and compact deep
  neural networks.
\newblock In {\em {ECCV}}, volume 11212 of {\em Lecture Notes in Computer
  Science}, 2018, pp. 373--390.
\newblock DOI: 10.1007/978-3-030-01237-3\_23.

\bibitem{ABC-Net}
Lin X, Zhao C, Pan W.
\newblock Towards accurate binary convolutional neural network.
\newblock In {\em {NIPS}}, 2017, pp. 345--353.

\bibitem{AutoQ}
Lou Q, Guo F, Kim M, Liu L, Jiang L.
\newblock Autoq: Automated kernel-wise neural network quantization.
\newblock {\em ICLR}, 2020.

\bibitem{yang2021bsq}
Yang H, Duan L, Chen Y, Li H.
\newblock Bsq: Exploring bit-level sparsity for mixed-precision neural network
  quantization.
\newblock In {\em {ICLR}}, 2021.

\bibitem{gumbel_softmax1}
Jang E, Gu S, Poole B.
\newblock Categorical reparameterization with gumbel-softmax.
\newblock {\em arXiv}, 2016.

\bibitem{gumbel_softmax2}
Maddison C~J, Mnih A, Teh Y~W.
\newblock The concrete distribution: A continuous relaxation of discrete random
  variables.
\newblock {\em arXiv}, 2016.

\bibitem{QAT}
Jacob B, Kligys S, Chen B, Zhu M, Tang M, Howard A, Adam H, Kalenichenko D.
\newblock Quantization and training of neural networks for efficient
  integer-arithmetic-only inference.
\newblock In {\em {IEEE CVPR}}, 2018, pp. 2704--2713.
\newblock DOI: 10.1109/cvpr.2018.00286.

\bibitem{nair2010rectified}
Nair V, Hinton G~E.
\newblock Rectified linear units improve restricted boltzmann machines.
\newblock In {\em {ICML}}, 2010, pp. 807--814.

\bibitem{deng2009imagenet}
Deng J, Dong W, Socher R, Li L, Li K, Li F.
\newblock Imagenet: {A} large-scale hierarchical image database.
\newblock In {\em {IEEE CVPR}}, 2009, pp. 248--255.
\newblock DOI: 10.1109/cvpr.2009.5206848.

\bibitem{alexnet}
Krizhevsky A, Sutskever I, Hinton G~E.
\newblock Imagenet classification with deep convolutional neural networks.
\newblock In {\em {NIPS}}, 2012, pp. 1106--1114.

\bibitem{mobilenet}
Howard A~G, Zhu M, Chen B, Kalenichenko D, Wang W, Weyand T, Andreetto M, Adam
  H.
\newblock Mobilenets: Efficient convolutional neural networks for mobile vision
  applications.
\newblock {\em CoRR}, 2017, abs/1704.04861.

\bibitem{sandler2018mobilenetv2}
Sandler M, Howard A~G, Zhu M, Zhmoginov A, Chen L.
\newblock Mobilenetv2: Inverted residuals and linear bottlenecks.
\newblock In {\em {IEEE CVPR}}, 2018, pp. 4510--4520.
\newblock DOI: 10.1109/cvpr.2018.00474.

\bibitem{inceptionv3}
Szegedy C, Vanhoucke V, Ioffe S, Shlens J, Wojna Z.
\newblock Rethinking the inception architecture for computer vision.
\newblock In {\em {IEEE CVPR}}, 2016, pp. 2818--2826.
\newblock DOI: 10.1109/CVPR.2016.308.

\bibitem{simon2016cnnmodels}
Simon M, Rodner E, Denzler J.
\newblock Imagenet pre-trained models with batch normalization.
\newblock {\em CoRR}, 2016, abs/1612.01452.

\bibitem{gross2016training}
Gross S, Wilber M.
\newblock Training and investigating residual nets.
\newblock {\em Facebook AI Research}, 2016, 6(3).

\bibitem{DSQ}
Gong R, Liu X, Jiang S, Li T, Hu P, Lin J, Yu F, Yan J.
\newblock Differentiable soft quantization: Bridging full-precision and low-bit
  neural networks.
\newblock In {\em {IEEE ICCV}}, 2019, pp. 4851--4860.
\newblock DOI: 10.1109/iccv.2019.00495.

\bibitem{BCGD}
Yin P, Zhang S, Lyu J, Osher S~J, Qi Y, Xin J.
\newblock Blended coarse gradient descent for full quantization of deep neural
  networks.
\newblock {\em CoRR}, 2018, abs/1808.05240.

\bibitem{he2016effectivequantization}
He Q, Wen H, Zhou S, Wu Y, Yao C, Zhou X, Zou Y.
\newblock Effective quantization methods for recurrent neural networks.
\newblock {\em arXiv preprint arXiv:1611.10176}, 2016.

\bibitem{hubara2017quantizednn}
Hubara I, Courbariaux M, Soudry D, El-Yaniv R, Bengio Y.
\newblock Quantized neural networks: Training neural networks with low
  precision weights and activations.
\newblock {\em The Journal of Machine Learning Research}, 2017,
  18(1):6869--6898.

\bibitem{kapur2017low}
Kapur S, Mishra A, Marr D.
\newblock Low precision rnns: Quantizing rnns without losing accuracy.
\newblock {\em arXiv preprint arXiv:1710.07706}, 2017.

\bibitem{zhou2017balanced}
Zhou S~C, Wang Y~Z, Wen H, He Q~Y, Zou Y~H.
\newblock Balanced quantization: An effective and efficient approach to
  quantized neural networks.
\newblock {\em Journal of Computer Science and Technology}, 2017,
  32(4):667--682.
\newblock DOI: 10.1007/s11390-017-1750-y.

\bibitem{peiqi2018hitnet}
Wang P, Xie X, Deng L, Li G, Wang D, Xie Y.
\newblock Hitnet: Hybrid ternary recurrent neural network.
\newblock In {\em {NIPS}}, volume~31, 2018.

\bibitem{hochreiter1997lstm}
Hochreiter S, Schmidhuber J.
\newblock Long short-term memory.
\newblock {\em Neural computation}, 1997, 9(8):1735--1780.

\bibitem{taylor2003ptb}
Taylor A, Marcus M, Santorini B.
\newblock The penn treebank: an overview.
\newblock {\em Treebanks}, 2003, pp. 5--22.
\newblock DOI: 10.1007/978-94-010-0201-1\_1.

\bibitem{kingma2015adam}
Kingma D~P, Ba J.
\newblock Adam: A method for stochastic optimization.
\newblock In {\em {ICLR}}, 2015.

\bibitem{krizhevsky2009learning}
Krizhevsky A, Hinton G.
\newblock Learning multiple layers of features from tiny images.
\newblock Technical report, Citeseer, 2009.

\bibitem{simonyan2014vgg}
Simonyan K, Zisserman A.
\newblock Very deep convolutional networks for large-scale image recognition.
\newblock {\em ICLR}, 2015.

\bibitem{lee2015deeply}
Lee C, Xie S, Gallagher P~W, Zhang Z, Tu Z.
\newblock Deeply-supervised nets.
\newblock In {\em AISTATS}, volume~38, 2015.

\end{thebibliography}

\vspace{5mm}

\noindent\parbox{8.3cm}{\parpic{\includegraphics[width=1in,height=1.25in,clip]{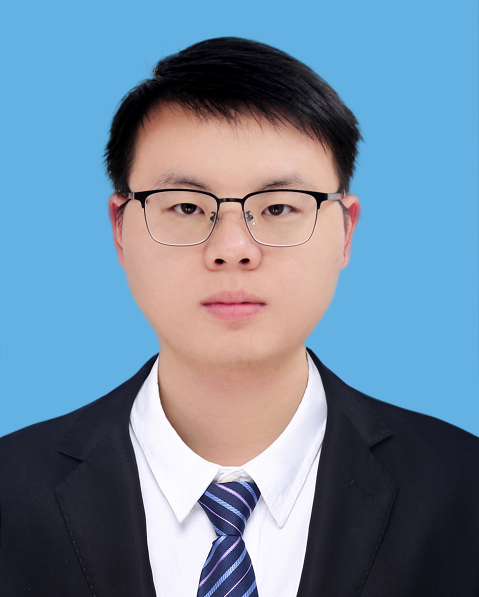}}{\small\quad {\bf Cheng Gong} received his B.S. and Ph.D. degrees in computer science and technology from Nankai University, Tianjin, in 2016 and 2022, respectively. 
He is a postdoctoral fellow at the College of Software, Nankai University, Tianjin.
His main research interests include neural network compression, heterogeneous computing, and machine learning.}\\[1mm]}

\noindent\parbox{8.3cm}{\parpic{\includegraphics[width=1in,height=1.25in,clip]{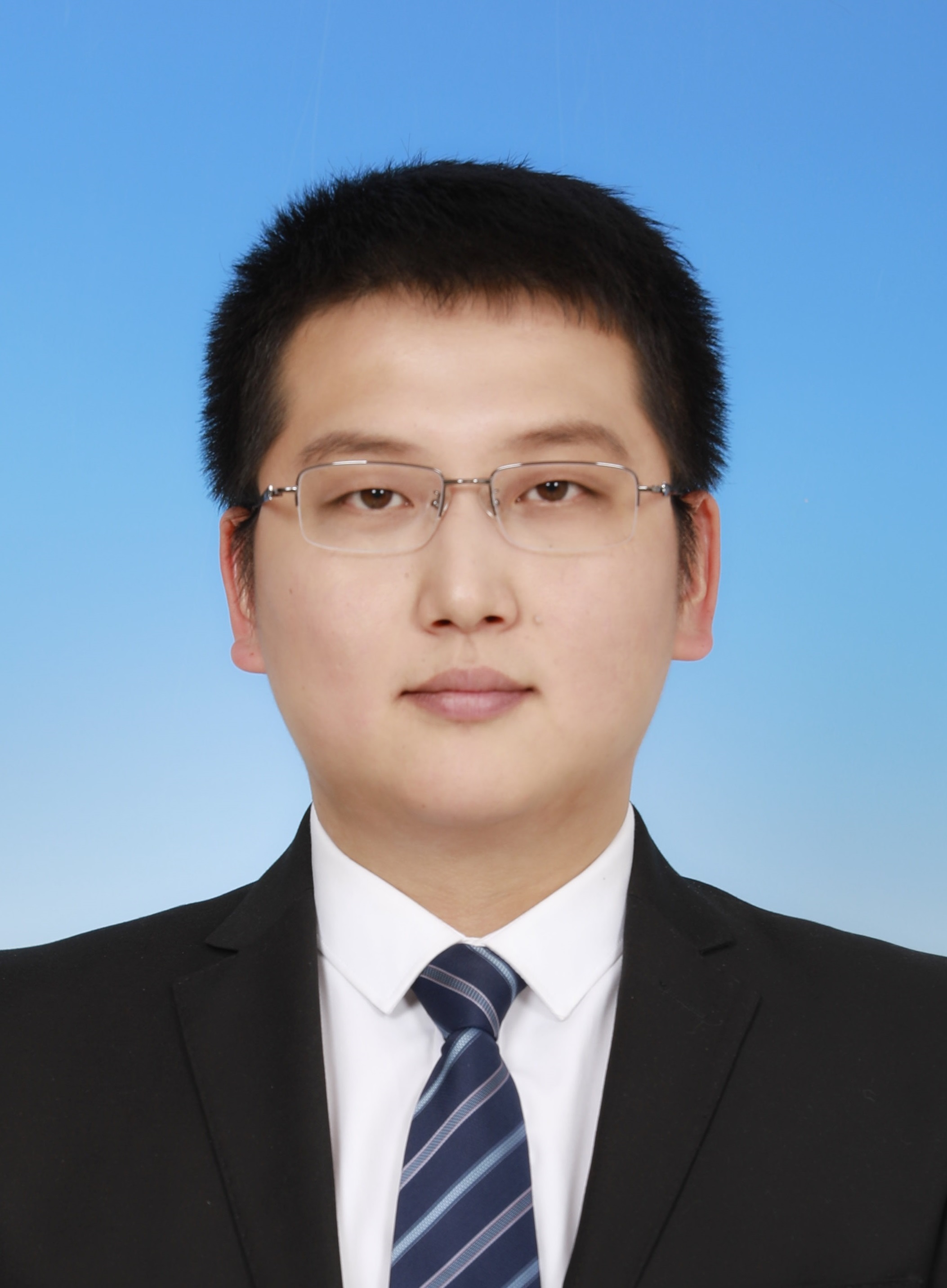}}{\small\quad {\bf Ye Lu} received his B.S. and Ph.D. degrees from Nankai University, Tianjin, in 2010 and 2015, respectively. 
He is an associate professor at the College of Cyber Science, Nankai University, Tianjin.
His main research interests include DNN FPGA accelerator, blockchian virtual machine, embedded system, and Internet of Things. }\\[1mm]}

\noindent\parbox{8.3cm}{\parpic{\includegraphics[width=1in,height=1.25in,clip]{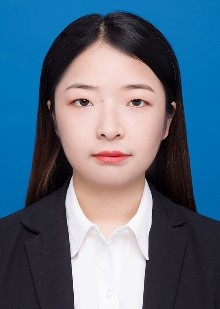}}{\small\quad {\bf Surong Dai} received her B.S. degree in computer science and technology from Nankai University, Tianjin, in 2020. 
She is currently working toward her Ph.D. degree in the College of Computer Science, Nankai University, Tianjin. 
Her main research interests include computer architecture, compiler design, and blockchain virtual machine.}\\[1mm]}

\noindent\parbox{8.3cm}{\parpic{\includegraphics[width=1in,height=1.25in,clip]{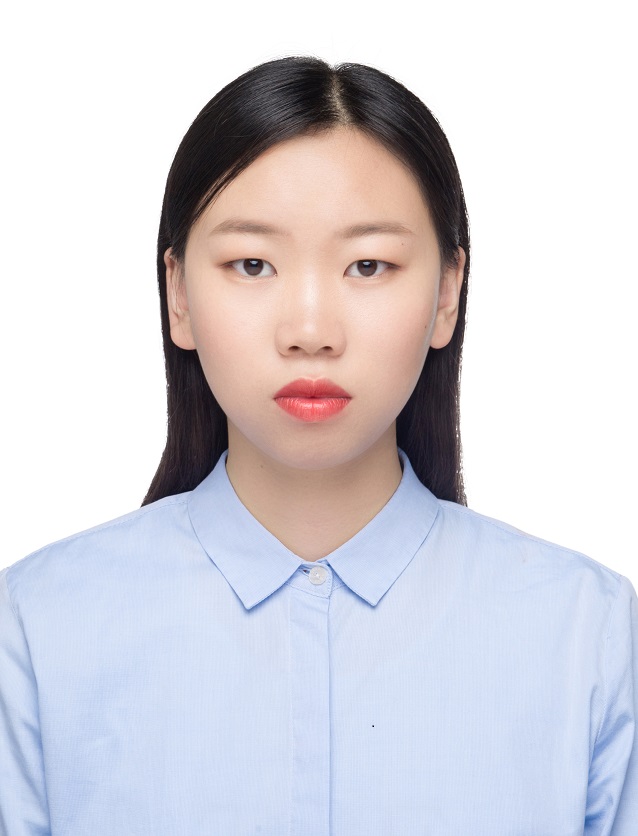}}{\small\quad {\bf Qian Deng} received her B.S. degree in computer science and technology from Nankai University in 2020.
She is currently working toward her M.S. degree in the College of Cyber Science, Nankai University, Tianjin.
Her main research interests include computer vision and artificial intelligence.}\\[1mm]}

\noindent\parbox{8.3cm}{\parpic{\includegraphics[width=1in,height=1.25in,clip]{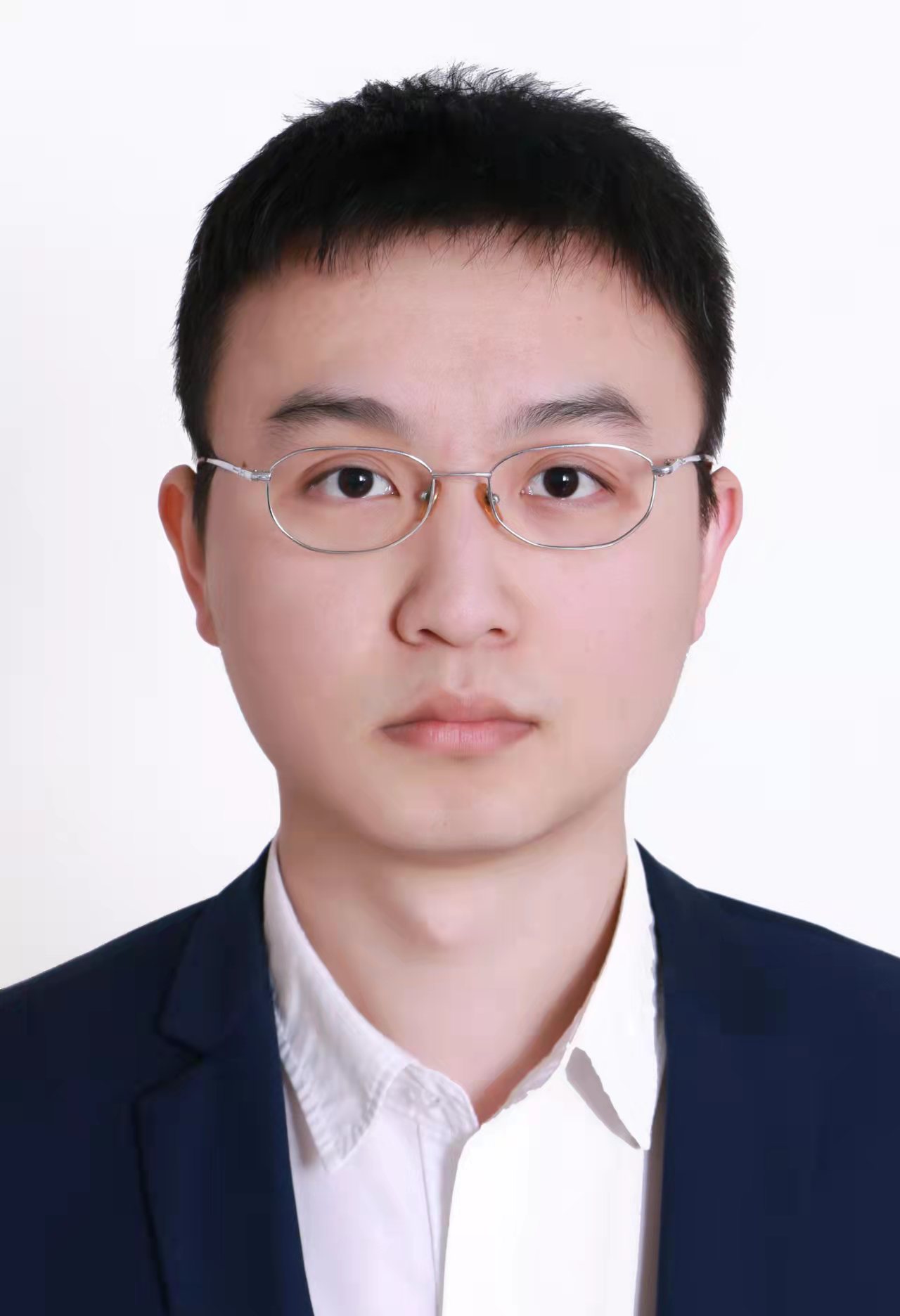}}{\small\quad {\bf Chengkun Du} received his B.S. degree from Nankai University in 2020. 
He is currently working toward his M.S. degree in the College of Computer Science, Nankai University, Tianjin. 
His main research interests include heterogeneous computing and machine learning.}\\[1mm]}

\noindent\parbox{8.3cm}{\parpic{\includegraphics[width=1in,height=1.25in,clip]{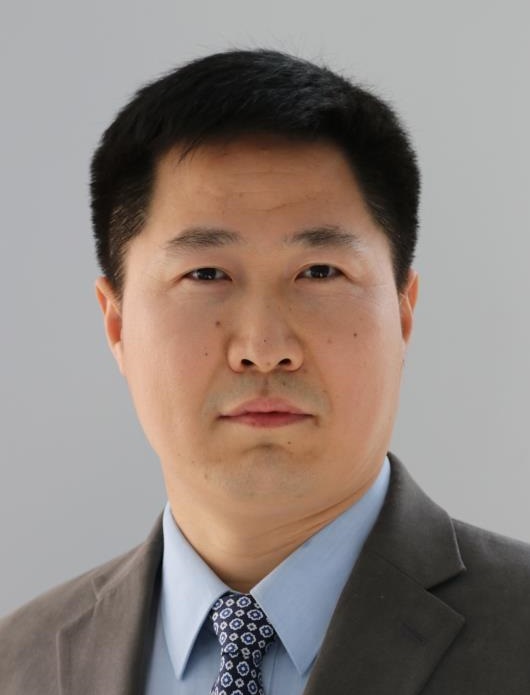}}{\small\quad {\bf Tao Li} received his Ph.D. degree in Computer Science from Nankai University, in 2007. 
He works at the College of Computer Science, Nankai University as a professor. 
He is the member of the IEEE Computer Society and the ACM, and the distinguished member of the CCF. 
His main research interests include heterogeneous computing, machine learning, and blockchain system. }\\[1mm]}

\label{last-page}
\end{multicols}
\end{document}


\begin{CJK*}{UTF8}{gbsn}
\thispagestyle{empty}
\vspace*{-13mm}
\noindent {\small Journal of computer science and technology: Instruction for authors.
JOURNAL OF COMPUTER SCIENCE AND TECHNOLOGY}
\vspace*{2mm}

\centering{\Large{\textbf{Supplementary Materials}}}




\end{CJK*}
\baselineskip=18pt plus.2pt minus.2pt
\parskip=0pt plus.2pt minus0.2pt
\begin{multicols}{2}
\section{Salient Object Detection}

\begin{table*}[ht]
\caption{\label{tab:comparing_results_on_sod}\gccpf{Evaluation Results of the Salient Object Detection Models on Four Metrics}}  
\setlength{\tabcolsep}{3pt}
\begin{tabular}{lccccccccccccc}\Xhline{1.5pt}
{Models} & {W/A} & \multicolumn{4}{c}{MSRA-B~\cite{msra}} & \multicolumn{4}{c}{DUTs~\cite{duts}} & \multicolumn{4}{c}{DUT-OMRON~\cite{duts}} \\\cmidrule(r){3-6}  \cmidrule(r){7-10} \cmidrule(r){11-14}
 &  & IoU$\uparrow$ & F$_1$$\uparrow$ & EM$\uparrow$ & MAE$\downarrow$ & IoU$\uparrow$ & F$_1$$\uparrow$ & EM$\uparrow$ & MAE$\downarrow$ & IoU$\uparrow$& F$_1$$\uparrow$ & EM$\uparrow$ & MAE$\downarrow$ \\\Xhline{.5pt}
{UNet} & 32/32 & 0.8936 & 0.9437 & 0.9645 & 0.0244 & 0.6106 & 0.7575 & 0.8417 & 0.0773 & 0.6390 & 0.7779 & 0.8262 & 0.0696 \\
 & 2.15/2.76 & 0.8850 & 0.9389 & 0.9616 & 0.0265 & 0.6074 & 0.7549 & 0.8358 & 0.0775 & 0.6390 & 0.7779 & 0.8236 & 0.0688 \\
{FPN} & 32/32 &  0.8920 & 0.9428 & 0.9644 & 0.0248 & 0.6286 & 0.7713 & 0.8505 & 0.0737 & 0.6316 & 0.7729 & 0.8292 & 0.0732 \\
 & 2.20/2.99 & 0.8908 & 0.9422 & 0.9644 & 0.0251 & 0.6298 & 0.7720 & 0.8469 & 0.0719 & 0.6495 & 0.7865 & 0.8346 & 0.0662 \\
{LinkNet} & 32/32 & 0.8904 & 0.9419 & 0.9640 & 0.0252 & 0.6286 & 0.7713 & 0.8519 & 0.0725 & 0.6426 & 0.7809 & 0.8386 & 0.0690 \\
 & 2.20/2.65 &  0.8858 & 0.9393 & 0.9619 & 0.0262 & 0.6234 & 0.7673 & 0.8425 & 0.0726 & 0.6438 & 0.7821 & 0.8272 & 0.0673 \\
\Xhline{1.5pt}
\end{tabular}\flushleft\vspace{-8pt}
\gccpf{Note: }{The $\uparrow$ denotes that the higher values shows better results and the $\downarrow$ vise versa.} 
\gccpf{W/A idicates the average bitwidth for weights and activation, respectively.}
\end{table*}
We employ MSRA-10K~\cite{MSRA10K} as the training set, and take MSRA-B~\cite{msra}, DUTs~\cite{duts}, and DUT-OMRON~\cite{duts} as the validation set.
All the images in these datasets are resized to $224\times 224$ for the training and test.
Three models including UNet~\cite{ronneberger2015unet}, LinkNet~\cite{chaurasia2017linknet}, and FPN~\cite{lin2017fpn} are involved in the evaluation.
We take ResNet50~\cite{he2016resnet} as the backbone network for these models and the implementation details can be found here\footnote{https://github.com/qubvel/segmentation\_models, \text{Oct. 2022}.\label{segementation_model_note}}.
Besides, we take the intersection over union (IoU), F$_1$-measure (F$_1$), E-measure (EM)~\cite{fan2018enhanced}, and mean absolute error (MAE) as the evaluation metrics.
The implementations of IoU, F$_1$ and MAE can be found here\textsuperscript{\ref{segementation_model_note}}, and EM refers to \cite{fan2018enhanced}.
We employ ClipQ for model quantization and do not quantize the first and last layers of models.
\gcc{We set the average bitwidth as $2$ for QPL to apply an extremely low-bitwidth quantization for these large SOD models for great computing benefits.}

The quantitative comparison results are in \tabref{tab:comparing_results_on_sod}. 
The model size and memory footprint of UNet are reduced by \text{14.88}$\times$ and \text{172.56}$\times$, respectively, when it is quantized into the average bitwidths of 2.15/2.76 with AutoQNN, and the performance degradation in four metrics across all datasets is less than \text{0.01}. 
AutoQNN also shrinks the memory footprint of FPN and LinkNet by \text{155.67}$\times$ and \text{172.64}$\times$ through quantizing models from full precision into low precision with average bitwidths of 2.20/2.99 and 2.20/2.65, respectively, and only incurs slight performance degradation which is not more than \text{0.018} in all metrics across three datasets.
The results demonstrate that QPL can reduce memory footprint by using the extremely low-bitwidth weight and activation with negligible performance degradation. 
This highlights that QPL can search for a suitable quantizing strategy to maintain comparable feature extraction quality of SOD models. 
\vspace{2mm}
\bibliographystyle{JCST}
\bibliography{references}

\label{last-page}
\end{multicols}